\pdfoutput=1

\documentclass[10pt,twocolumn,letterpaper]{article}

\usepackage[pagenumbers]{cvpr}      %

\usepackage{multirow}
\usepackage{array}
\usepackage{microtype}
\usepackage{colortbl}
\usepackage{tabularx}
\usepackage{placeins}
\usepackage{capt-of}
\definecolor{up}{HTML}{2E8B57}
\definecolor{down}{HTML}{C44E52}
\definecolor{bench}{HTML}{4C72B0}

\definecolor{cvprblue}{rgb}{0.21,0.49,0.74}
\usepackage[pagebackref,breaklinks,colorlinks,allcolors=cvprblue]{hyperref}

\title{VisionQ: Benchmarking VLM Judges on Criterion-Specific Comparisons\\of Computer Vision Method Outputs}

\author{%
Vu Dinh Xuan$^{1*}$ \quad Duc-Hai Nguyen$^{2*}$ \quad Minh-Dung Dao$^{2}$ \quad Vu Quynh Giao$^{2}$\\
Quang Hong Nguyen$^{3}$ \quad Binh-Son Hua$^{4}$ \quad Barry O'Sullivan$^{2}$\\
David Murphy$^{2}$ \quad Hoang D. Nguyen$^{2\dagger}$\\[4pt]
{\small $^{1}$University of Information Technology, VNU-HCM, Vietnam \quad $^{2}$University College Cork, Ireland}\\
{\small $^{3}$Hanoi University of Science and Technology, Vietnam \quad $^{4}$Trinity College Dublin, Ireland}\\
{\small $^{*}$Equal contribution \quad $^{\dagger}$Corresponding author: \texttt{hn@cs.ucc.ie}}
}

\begin{document}
\maketitle

\begin{abstract}
Qualitative comparison figures in computer vision papers are commonly established to illustrate which method's output is best on a named criterion, such as visual sharpness or surface complexity.
VisionQ investigates this important part of computer vision work beyond human judgment towards objective vision-language models (VLMs) as judges.
From 1{,}409 CVPR and ICCV papers, it creates blind multiple-choice questions, each testing one criterion, with the correct answer based on the authors' claimed best result.
The best of 21 judges aligns with the authors' claims on 69\% of questions, against 32\% by chance.
Yet it keeps most of its accuracy without the criterion and changes its answers when the options are reordered.
Trained on every option order of a larger question set, our VisionQ-Judge raises accuracy on unseen papers from 50\% to 61\%.
The paper highlights the potential of VLM judges to provide objective, criterion-specific comparisons across a broad range of computer vision methods in the past, present, and future.
Data, weights and code are available at \href{https://huggingface.co/visionq-review}{huggingface.co/visionq-review} and \href{https://github.com/ReML-AI/visionq}{github.com/ReML-AI/visionq}.
\end{abstract}

\section{Introduction}
\label{sec:intro}

\begin{figure*}[t]
\centering
\includegraphics[width=\textwidth]{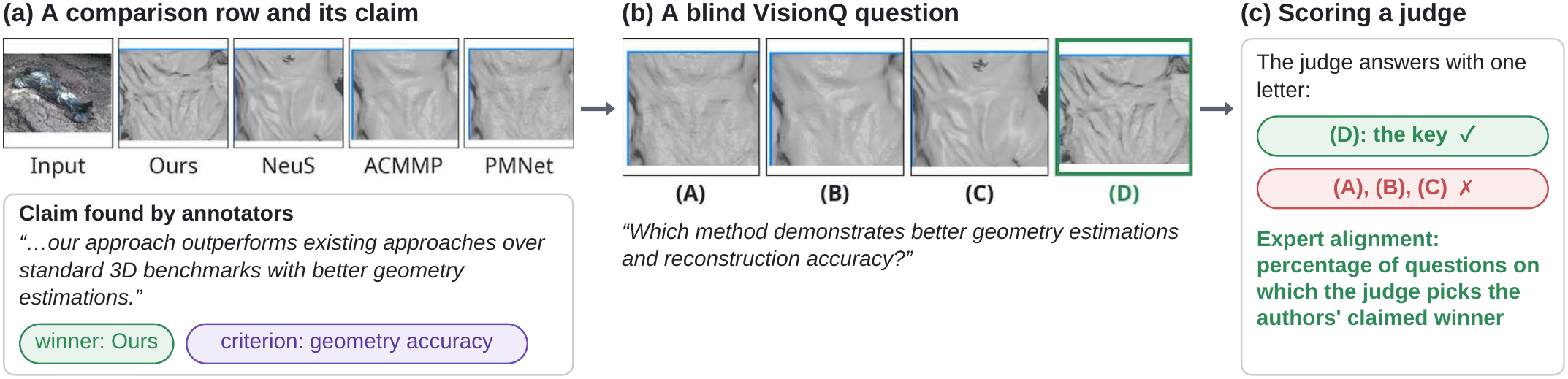}
\caption{VisionQ asks a judge which output is best on a named criterion. It scores the judge's choice against the source authors' claimed winner. (a)~For each comparison row, annotators find the authors' claim naming its winner and criterion. (b)~Crops from the row form a blind question with lettered options, hiding method names, caption and claim. The authors' claimed winner, framed in green here only, is the answer key. (c)~Expert alignment is the percentage of questions on which the judge chooses the authors' claimed winner when the question names the claimed criterion. \Cref{fig:construction} shows how the questions are built, and \cref{fig:teaser} shows how judges answer this one.}
\label{fig:overview}
\end{figure*}

Computer vision now publishes new methods at a pace that makes careful human checking difficult: CVPR grew from 1{,}660 papers in 2021 to 2{,}871 in 2025~\cite{cvfopenaccess}.
A new method typically earns its place by showing its outputs next to those of prior work in a qualitative comparison figure.
The text around each figure states which output is best on a named criterion, for example which one preserves thin structures, completes the surface or follows the prompt.
These claims are usually checked by human readers.

Checking them is a natural task for vision-language models (VLMs), which already act as judges of visual quality and of conditional generation~\cite{qbench,viescore,llavacritic,imagereward,hpsv2}.
A VLM judge built for these comparisons could compare the outputs of past, present and future methods objectively on any named criterion.
Before such a judge can take over, the field needs to know how often it agrees with a careful reader and what this agreement depends on.
A judge that responds to something other than the stated criterion would change which method looks best.

Answering these questions requires a benchmark with (a) outputs of competing methods on the same input, (b) questions that name the criterion, (c) a key whose provenance is documented and (d) checks of what the judge's agreement rests on.
Existing benchmarks for visual comparison ask about named attributes of natural, distorted or generated images, with answer keys written by expert or crowd annotators~\cite{qbenchplus,coinstruct,abench,mllmcompbench}.
The outputs of CV methods are a different kind of evidence: depth maps, meshes, masks and edited photographs, placed side by side by authors who expect them to differ on one criterion.
These benchmarks meet (b) at least in part, but not (a) (\cref{tab:related}).
To address these limitations, we introduce VisionQ, a benchmark dataset built upon the results reported in computer vision papers.
The dataset satisfies all four requirements (a)--(d), and its answer keys come from the claims that computer vision authors make (\cref{fig:overview}).

Keying a benchmark to author claims makes every answer traceable to a published statement, but the claimed winner is likely the source paper's own method or a variant of it.
Agreement with this key could therefore come from a preference for polished outputs or from cues outside the images rather than from the named criterion.
Option position is a further risk, because comparison figures often place the authors' method in the last column and judges are known to favour certain positions~\cite{mllmasajudge,wang2023large}.
To remove this cue, VisionQ shuffles the options, so the key is last on 32.0\% of VisionQ-Bench questions, close to the 32.2\% expected by chance.
A position habit then cannot raise a judge's score, although it can still change the judge's answers.
We test these alternatives with four input controls.
Because the key reflects the authors' claims about their own method, a blinded human study tests whether independent raters choose the same winner from the input a judge receives.

We evaluate 21 VLM judges from ten model families on VisionQ-Bench, ranging from 4B open-weight models to commercial models such as GPT-5.5 and Gemini 3.1 Pro.
Each judge is compared with chance and with image-quality baselines.
The strongest, Gemini 3.1 Pro, reaches an expert alignment of 68.6\%, roughly twice the 32.2\% chance level and above the 41.7\% of the best no-reference image-quality model.
Our analysis yields three findings.
(i) Blinded raters support the key. Judges agree with it more often where both raters pick it. Even there, the strongest judge disagrees with it on about one question in five.
(ii) For the six judges we control, agreement depends on the images but hardly on the named criterion: blank images bring them near chance, whereas asking which output is better overall keeps 86--97\% of their accuracy.
(iii) These judges also change their answers when the options are reordered.
Current judges therefore align with the authors' claims well above chance, but their choices rest on the images far more than on the named criterion, so they are not yet criterion-specific judges.

None of these judges was trained for this task, so we also train our own judge, VisionQ-Judge, from Gemma-4-E4B with DPO on preference pairs that differ only in the chosen letter.
We train it on VisionQ-MCQ, a larger question set from the same corpus, with the options shown in every cyclic order.
On held-out test questions from unseen papers, VisionQ-Judge raises the accuracy of its base model from 50.4\% to 61.0\%.

The three findings above matter to anyone who uses a VLM judge to compare methods.
A high score shows that a judge agrees with the authors, but it does not show that the judge followed the named criterion.
Before its verdicts are trusted, such a judge should pass criterion and option-order controls like ours.
Together, these results measure how far current VLM judges are from objective, criterion-specific comparison of computer vision methods, and VisionQ-Judge takes a first step towards closing that distance. Our contributions are:
\begin{itemize}
\item \textbf{VisionQ-Bench}: from 3{,}911 author-claimed comparisons in a corpus of 1{,}409 CVPR and ICCV papers, we distil 309 blind multiple-choice questions from 109 papers. Every claim is labelled by two annotators with adjudication, and the questions pass a rule-based filter. Each question names one criterion from a six-axis vocabulary, and VisionQ-MCQ adds 4{,}524 questions from the same pool to train and test judges. Together they meet (a)--(c) (\cref{sec:visionq}).
\item \textbf{A validity study} for (d): a blinded human study, image-quality baselines and four input controls that show what judge agreement rests on (\cref{sec:validity,sec:results}).
\item \textbf{VisionQ-Judge}, a first step towards such a judge: Gemma-4-E4B tuned with DPO~\cite{dpo} on every cyclic order of the options, which gains 10.6 points and lowers its last-option bias (\cref{sec:judge}).
\end{itemize}

\section{Related work}
\label{sec:related}

\begin{table}[t]
\centering
\caption{VisionQ and the closest benchmarks. \emph{Crit.}: questions name a criterion; \emph{Methods}: options are outputs of competing methods on one input; \emph{Human}: human baseline or agreement study; \emph{Order}: option-order control. \checkmark{} yes, $\circ$ partly, -- no or not reported. $^a$Blinded human study; $^b$option-order control (\cref{sec:validity}).}
\label{tab:related}
\scriptsize
\setlength{\tabcolsep}{1.1pt}
\begin{tabularx}{\linewidth}{@{}l>{\raggedright\arraybackslash}Xcccc@{}}
\toprule
\textbf{Benchmark} & \textbf{Judged items} & \textbf{Crit.} & \textbf{Methods} & \textbf{Human} & \textbf{Order} \\
\midrule
Q-Bench+~\cite{qbenchplus} & photos, pairs & $\circ$ & -- & \checkmark & $\circ$ \\
MICBench~\cite{coinstruct} & 3--4 images & $\circ$ & -- & \checkmark & -- \\
A-Bench~\cite{abench} & generated images & $\circ$ & -- & \checkmark & $\circ$ \\
CompBench~\cite{mllmcompbench} & photo pairs & $\circ$ & -- & \checkmark & -- \\
SubtleBench~\cite{vlmsubtlebench} & image pairs & \checkmark & -- & \checkmark & $\circ$ \\
MJ-Bench~\cite{mjbench} & generated images & \checkmark & $\circ$ & $\circ$ & \checkmark \\
VL-RewardBench~\cite{vlrewardbench} & text responses & -- & $\circ$ & $\circ$ & \checkmark \\
\midrule
\rowcolor{bench!12}\textbf{VisionQ} & method outputs & \checkmark & \checkmark & \checkmark$^a$ & \checkmark$^b$ \\
\bottomrule
\end{tabularx}
\end{table}

\paragraph{Attribute-specific visual comparison.}
Benchmarks for low-level perception ask VLMs about named attributes of natural~\cite{qbench} and generated images~\cite{abench}, and later ones extend this format to pairs and groups of images~\cite{qbenchplus,coinstruct,mllmcompbench,vlmsubtlebench}.
Their images are photographs, generated images or domain images, and their answer keys come from annotators or dataset labels.
VisionQ keeps the question format but changes both: its options are outputs of competing CV methods on the same input, and its key is the claim that the source paper makes about them (\cref{tab:related}).

\paragraph{VLMs as judges and their biases.}
Meta-evaluations of VLM judges measure agreement with human preferences on responses, generated images or image edits~\cite{mllmasajudge,vlrewardbench,mjbench,mllmjudgeediting}, or with expert ratings, as VisJudge-Bench~\cite{visjudge} does for visualisations.
In VisionQ, the experts are the authors of the compared methods.
These and other studies document judge biases towards option positions and longer responses~\cite{mllmasajudge,vlrewardbench,mjbench,mmjudgebias}.
Zheng et al.~\cite{wang2023large} showed that LLMs prefer certain option positions and reduced this preference at inference time with PriDe, which \cref{sec:judge} compares with VisionQ-Judge.
To expose such biases, these studies reorder the options~\cite{vlrewardbench,mjbench} or remove or perturb the image~\cite{mllmasajudge,mmjudgebias}; VisionQ applies both controls to test what agreement with its key rests on.

\paragraph{Claims in scientific figures.}
Multimodal claim verification asks whether a figure or table supports a claim from a paper: MuSciClaims~\cite{musciclaims} perturbs the claims and SciClaimEval~\cite{sciclaimeval} edits the figures and tables to build contradicted examples.
VisionQ also starts from claims, but asks the judge to choose among the method outputs that a claim compares, as a reader does.

\paragraph{Training judges.}
Judges have been trained with supervised critiques~\cite{llavacritic}, reinforcement learning on quality rankings~\cite{visualqualityr1} and DPO on contrastive judgments~\cite{conj}.
VisionQ-Judge also uses DPO but builds its preference pairs differently: the two responses share one evidence sentence and differ only in the chosen letter, and the pairs are built for every cyclic order of the options.

\section{VisionQ}
\label{sec:visionq}

\begin{figure*}[t]
\centering
\includegraphics[width=\textwidth]{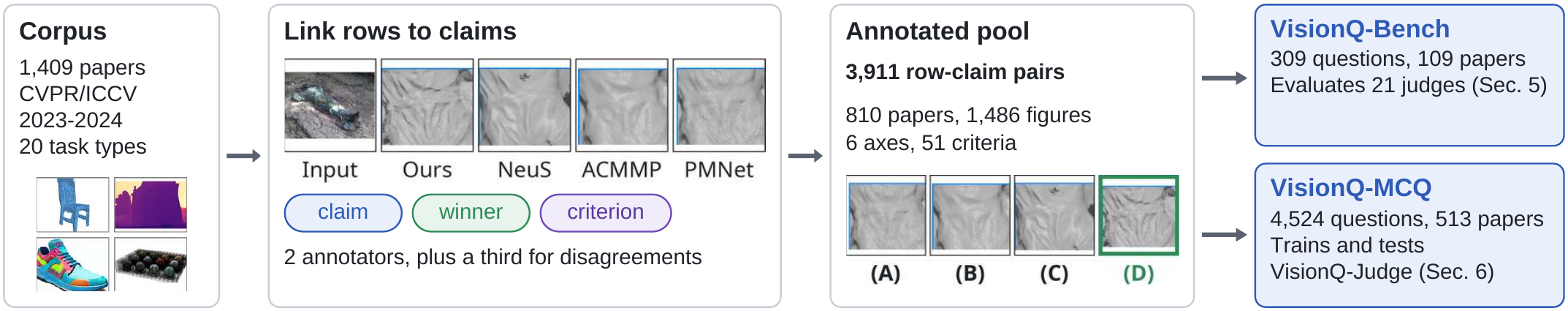}
\caption{Construction of VisionQ. The answer key comes from the authors' claim, not from a new judgment of the images. Annotators only link each comparison row to the claim naming its winner and criterion. The annotated pool provides VisionQ-Bench for evaluating judges and VisionQ-MCQ for training and testing them. Fig.~S1 shows the full pipeline.}
\label{fig:construction}
\end{figure*}

This section defines a VisionQ question, its source papers, its answer key and its criteria (\cref{sec:task,sec:items,sec:key,sec:vocab}) and then describes the three parts built from them: VisionQ-Bench, VisionQ-MCQ and VisionQ-Judge (\cref{sec:release}).

\subsection{Task and inputs}
\label{sec:task}

A VisionQ question pairs one composite image with one sentence (\cref{fig:overview}).
The composite holds two to four crops of method outputs from one comparison row, each under a rendered letter, and the sentence asks which output is better on a named criterion.
Method labels, the caption and the author claim are withheld.
A judge answers with one letter, and unparseable responses count as wrong.
The questions mix two, three and four options, so scores are read against an item-weighted chance level of 32.2\%, and all intervals are 95\% intervals from a bootstrap over source papers (App.~\ref{sec:s-protocol}).

\subsection{Sources and filtering}
\label{sec:items}

The corpus holds 1{,}409 papers from CVPR 2023, ICCV 2023 and CVPR 2024 across 20 task types; 810 of them supplied annotated figures, yielding 3{,}911 data points, each a comparison row linked to one claim (\cref{fig:construction}; App.~\ref{sec:s-items}).
VisionQ-Bench draws 332 questions from 114 of these papers.
A rule-based filter on option labels removes 23 that do not form a valid comparison, such as two options with the same method name, which leaves 309 questions from 109 papers.

\subsection{Answer key}
\label{sec:key}

Each answer key is an author claim.
We trust these claims for three reasons.
First, the authors know their task, data and failure cases best, and they produced the outputs themselves.
Second, each claim is published, so every key can be traced to a sentence in a peer-reviewed paper.
Third, they supply keys for thousands of comparisons, and annotators do not have to decide which output is better.
We therefore take the claimed winner as the correct answer but treat it as expert judgment rather than ground truth.
Annotation links a comparison row to the sentence in the paper that states which method is better on which criterion, and the claimed winner's crop becomes the correct option.
Two annotators label each item, a third resolves disagreements, and comparisons without a single winner are discarded.
The annotators record the claim; they do not choose the winner.

This procedure fixes who wins: the correct option is the source paper's method or a variant of it, since 235 winning crops are labelled ``ours'', 47 carry the method name from the paper's title and 27 another name for the method or a variant.
We therefore define the measured quantity narrowly.
\emph{Expert alignment} is agreement with the source-paper authors' claimed winner under a question naming the claimed criterion.
Because it is anchored in a published claim rather than a new annotation, it does not certify that the claim is correct; \cref{sec:validity} measures how close it comes to independent readers.

\subsection{Criterion vocabulary}
\label{sec:vocab}

The criteria come from a vocabulary of six axes and 51 leaves, built from 9{,}228 claims in the corpus by open coding and hierarchical clustering following Clio~\cite{tamkin2024clio}, with human review of every label (App.~\ref{sec:s-vocab}).
At the image level, Image Appearance covers low-level qualities such as blur, noise and colour, and Object Form covers shape, boundary, texture and completeness.
At the scene level, Scene Layout covers composition, depth and viewpoint, and Relation covers contact, correspondence and attribute binding.
Across images, Reference Fidelity covers correspondence with a source image or ground truth, and Prompt Match covers agreement with the conditioning prompt.
The vocabulary covers 96.0\% of the 1{,}486 annotated figures, and fields lean on different leaves: papers on 3D reconstruction place 57.8\% of their claims on three Object Form leaves, whereas papers on 2D generation place 30.2\% on one Prompt Match leaf.

\subsection{Released components}
\label{sec:release}

VisionQ has three parts, each with one purpose.
\emph{VisionQ-Bench}, the 309 questions of \cref{sec:items}, is used only for evaluation.
It scores the 21 judges and is also the question set of the human study and the input controls (\cref{sec:validity,sec:results}).
\emph{VisionQ-MCQ} holds 4{,}524 questions from 513 papers in the same corpus.
After a cap on questions per criterion leaf, it is split by paper into training, validation and test sets, so a trained judge is tested on unseen papers.
\emph{VisionQ-Judge} is Gemma-4-E4B tuned with DPO on this training split, with the options in every cyclic order (\cref{sec:judge}).
The criterion vocabulary, the source-paper list and the code are released with them.

\section{Validity study}
\label{sec:validity}

\Cref{sec:key} left two questions open.
First, would an independent reader who sees only the images and the question pick the authors' claimed winner?
Second, when a judge picks this winner, does it follow the named criterion or other cues?
A blinded human study answers the first question, and four input controls answer the second.

\paragraph{Blinded human study (H).}
The key records the authors' claims about their own method.
Two CV-literate raters answer 126 of the 309 questions blind to measure independent agreement with it.
These questions come from 87 papers and were chosen to cover every axis and as many papers as possible.
Each rater sees the same composite and full question as a judge, without method names, claims or the key.
Options appear in a private order, and raters may answer ``cannot tell''.
Raters are told to judge the stated criterion, not which image looks most attractive overall.
Where the source figure has a reference image, a second screen adds it and asks again.
The raters are not the annotators of \cref{sec:key}. Their answers do not filter or change the key.
Instead, the answers show how often raters who see only the judge's input choose the authors' winner.
The answers also mark questions where both raters pick the key and others where at least one picks another crop (App.~\ref{sec:s-human}).
Where both raters pick the key, a judge that chooses another crop disagrees with both the raters and the authors.
Where they do not, the crops may lack visual evidence, the raters may read the criterion differently, or the key may be debatable.

\paragraph{Input controls (E1--E4).}
A judge can match the key without following the named criterion.
E1--E3 therefore each change one input to test one of the alternative explanations given in \cref{sec:intro}, and E4 tests the standard protocol itself.
The controls run on six judges: the Gemma-4-E4B base model and five judges that span the leaderboard, from Gemma-4-26B-A4B (48.5\%) to Gemini 3.1 Pro (68.6\%).
Each control is reported as a paired difference from the standard condition on the same 309 questions (App.~\ref{sec:s-protocol}).
E1 tests the habit of picking one position: it renders every cyclic order of the options (1{,}029 composites) and reports the accuracy averaged over orders (\emph{mean over orders}), a \emph{plurality vote} over orders that counts ties as wrong, and the share of questions answered with one crop in every order (\emph{same crop}).
E2 tests the taste for polished outputs: it replaces the question with ``Which output is better overall?'', which should cost accuracy if the judge follows the criterion.
E3 tests cues outside the images: it replaces the crops with white tiles or with another question's composite, which should bring a judge that reads the images to chance.
E4 tests the standard protocol, which uses the released question strings, abbreviated for long criteria, and shows no reference image: it adds the source figure's reference row (205 questions) and restores the full question, with a text-only variant that separates the two changes (Appendices~\ref{sec:s-items} and~\ref{sec:s-protocol}).

\section{Results}
\label{sec:results}

\begin{table*}[t]
\centering
\caption{Expert alignment (\%) on VisionQ-Bench (309 questions; chance 32\%). Brackets: 95\% paper-bootstrap intervals. Bold: best in each column. $^\ddagger$Base model of VisionQ-Judge. Nine of the 21 judges are shown, and Tab.~S6 lists all of them with per-axis intervals. VisionQ-Judge is evaluated on VisionQ-MCQ (\cref{sec:judge}).}
\label{tab:leaderboard}
\small
\setlength{\tabcolsep}{4pt}
\begin{tabularx}{\textwidth}{@{}l*{1}{>{\centering\arraybackslash}X}@{\hspace{8pt}}*{3}{>{\centering\arraybackslash}X}@{\hspace{8pt}}*{3}{>{\centering\arraybackslash}X}@{}}
\toprule
\textbf{Judge} & \textbf{Overall} & \textbf{Object Form} & \textbf{Ref.\ Fidelity} & \textbf{Img.\ Appearance} & \textbf{Relation} & \textbf{Prompt Match} & \textbf{Scene Layout} \\
\midrule
Gemini 3.1 Pro & \textbf{69} {\scriptsize[62,74]} & \textbf{76} & 59 & \textbf{69} & 52 & \textbf{77} & 62 \\
GPT-5.3-codex & 63 {\scriptsize[57,69]} & 64 & 59 & \textbf{69} & \textbf{58} & 59 & \textbf{75} \\
GPT-5.5 & 63 {\scriptsize[57,69]} & 68 & \textbf{61} & 64 & 45 & 64 & 62 \\
MiMo v2.5 & 59 {\scriptsize[53,65]} & 64 & 52 & 67 & 45 & 59 & 50 \\
Claude Opus 4.6 & 58 {\scriptsize[52,65]} & 62 & \textbf{61} & 56 & 39 & 59 & 50 \\
Qwen 3.6-35B-A3B & 58 {\scriptsize[51,64]} & 62 & 48 & 62 & 45 & 64 & 62 \\
Gemma-4-31B & 57 {\scriptsize[51,64]} & 64 & 52 & 62 & 42 & 50 & 50 \\
Gemma-4-26B-A4B & 49 {\scriptsize[42,55]} & 54 & 39 & 59 & 32 & 45 & 50 \\
Gemma-4-E4B$^\ddagger$ & 45 {\scriptsize[39,51]} & 48 & 41 & 56 & 32 & 41 & 38 \\
\midrule
\emph{Chance} & \emph{32.2} & \emph{31.8} & \emph{32.8} & \emph{29.7} & \emph{34.9} & \emph{30.7} & \emph{40.6} \\
\bottomrule
\end{tabularx}
\end{table*}

\begin{table*}[t]
\centering
\caption{Input controls on VisionQ-Bench (309 questions). $\Delta$: paired change from the standard run (\colorbox{up!35}{green} up, \colorbox{down!35}{red} down; * 95\% paper-bootstrap interval excludes zero; --: n/a). Bold: best in the row. \emph{Same crop} at random: 13.8\%. $^\dagger$On the 205 questions with a reference image.}
\label{tab:controls}
\footnotesize
\setlength{\tabcolsep}{1pt}
\begin{tabularx}{\textwidth}{@{}l>{\raggedleft\arraybackslash}X>{\centering\arraybackslash}X>{\raggedleft\arraybackslash}X>{\centering\arraybackslash}X>{\raggedleft\arraybackslash}X>{\centering\arraybackslash}X>{\raggedleft\arraybackslash}X>{\centering\arraybackslash}X>{\raggedleft\arraybackslash}X>{\centering\arraybackslash}X>{\raggedleft\arraybackslash}X>{\centering\arraybackslash}X@{}}
\toprule
 & \multicolumn{2}{c}{\scriptsize\textbf{Gemma-4-E4B}} & \multicolumn{2}{c}{\scriptsize\textbf{Gemma-4-26B-A4B}} & \multicolumn{2}{c}{\scriptsize\textbf{Gemma-4-31B}} & \multicolumn{2}{c}{\scriptsize\textbf{MiMo v2.5}} & \multicolumn{2}{c}{\scriptsize\textbf{GPT-5.3-codex}} & \multicolumn{2}{c}{\scriptsize\textbf{Gemini 3.1 Pro}} \\
\cmidrule(lr){2-3}\cmidrule(lr){4-5}\cmidrule(lr){6-7}\cmidrule(lr){8-9}\cmidrule(lr){10-11}\cmidrule(lr){12-13}
Condition & Acc. & $\Delta$ & Acc. & $\Delta$ & Acc. & $\Delta$ & Acc. & $\Delta$ & Acc. & $\Delta$ & Acc. & $\Delta$ \\
\midrule
Standard & 45.6 & -- & 53.7 & -- & 56.0 & -- & 58.6 & -- & 61.2 & -- & \textbf{70.9} & -- \\
\midrule
\multicolumn{13}{@{}l}{\textbf{E1: option order}} \\
Mean over orders & 44.4 & \cellcolor{down!8}$-$1.2 & 54.1 & \cellcolor{up!7}+0.4 & 59.7 & \cellcolor{up!13}+3.7* & 59.8 & \cellcolor{up!8}+1.2 & 60.9 & \cellcolor{down!6}$-$0.2 & \textbf{69.2} & \cellcolor{down!9}$-$1.6 \\
Plurality vote & 36.9 & \cellcolor{down!23}$-$8.7* & 47.6 & \cellcolor{down!18}$-$6.1* & 60.2 & \cellcolor{up!14}+4.2* & 56.3 & \cellcolor{down!11}$-$2.3 & 57.6 & \cellcolor{down!13}$-$3.6 & \textbf{67.6} & \cellcolor{down!12}$-$3.2 \\
Same crop in all orders & 20.4 & -- & 41.1 & -- & 51.5 & -- & 44.3 & -- & 51.1 & -- & \textbf{64.7} & -- \\
\midrule
\multicolumn{13}{@{}l}{\textbf{E2: criterion}} \\
Criterion removed & 39.5 & \cellcolor{down!18}$-$6.1* & 52.1 & \cellcolor{down!9}$-$1.6 & 51.8 & \cellcolor{down!14}$-$4.2 & 55.3 & \cellcolor{down!12}$-$3.2 & 52.4 & \cellcolor{down!23}$-$8.7* & \textbf{63.8} & \cellcolor{down!20}$-$7.1* \\
\midrule
\multicolumn{13}{@{}l}{\textbf{E3: images}} \\
Blank tiles & 29.8 & \cellcolor{down!38}$-$15.9* & 30.1 & \cellcolor{down!53}$-$23.6* & 26.9 & \cellcolor{down!60}$-$29.1* & 27.8 & \cellcolor{down!60}$-$30.7* & 30.4 & \cellcolor{down!60}$-$30.7* & 31.1 & \cellcolor{down!60}$-$39.8* \\
Mismatched composite & 33.0 & \cellcolor{down!31}$-$12.6* & 31.4 & \cellcolor{down!51}$-$22.3* & 30.7 & \cellcolor{down!56}$-$25.2* & 27.5 & \cellcolor{down!60}$-$31.1* & 26.9 & \cellcolor{down!60}$-$34.3* & 30.4 & \cellcolor{down!60}$-$40.5* \\
\midrule
\multicolumn{13}{@{}l}{\textbf{E4: restored inputs}} \\
Full question & 45.3 & \cellcolor{down!7}$-$0.3 & 52.8 & \cellcolor{down!8}$-$1.0 & 58.6 & \cellcolor{up!11}+2.6 & 60.2 & \cellcolor{up!9}+1.6 & 61.8 & \cellcolor{up!7}+0.6 & \textbf{68.9} & \cellcolor{down!10}$-$1.9 \\
Full question + ref.$^\dagger$ & 37.1 & \cellcolor{down!24}$-$8.8* & 59.0 & \cellcolor{up!25}+9.3* & 63.4 & \cellcolor{up!23}+8.3* & 62.4 & \cellcolor{up!18}+5.9 & 69.3 & \cellcolor{up!26}+9.8* & \textbf{74.6} & \cellcolor{up!16}+4.9* \\
\bottomrule
\end{tabularx}
\end{table*}

\begin{table}[t]
\centering
\caption{Blinded human study on 126 questions from 87 papers (chance 33.9\%). Raters see the full question in a private option order and may answer ``cannot tell''. Brackets: 95\% paper-bootstrap intervals. \emph{Gap}: a judge's accuracy on questions where both raters choose the key, minus its accuracy where at least one chooses another crop. The remaining 19 questions have a ``cannot tell'' answer and no other crop.}
\label{tab:human}
\footnotesize
\setlength{\tabcolsep}{2.5pt}
\begin{tabular}{@{}lcc@{}}
\toprule
 & \textbf{Rater 1} & \textbf{Rater 2} \\
\midrule
Agree with the key & 57.9 {\scriptsize[49,67]} & 47.6 {\scriptsize[39,56]} \\
``Cannot tell'' & 0.8 & 30.2 \\
\textbf{Agree with the key when deciding} & \textbf{58.4} {\scriptsize[50,68]} & \textbf{68.2} {\scriptsize[58,78]} \\
Before $\rightarrow$ after the reference (79 q.) & 58.2 $\rightarrow$ \textcolor{up}{64.6} & 43.0 $\rightarrow$ \textcolor{up}{55.7} \\
Same crop when both decide (88 q.) & \multicolumn{2}{c}{70.5 {\scriptsize[61,80]}, $\kappa=0.59$} \\
\bottomrule
\end{tabular}

\vspace{4pt}
\begin{tabular}{@{}lcccc@{}}
\toprule
\textbf{Judge} & \shortstack{\textbf{Both pick}\\\textbf{key} (47)} & \shortstack{\\\textbf{Other} (60)} & \shortstack{\\\textbf{Gap}} & \shortstack{\\\textbf{All} (126)} \\
\midrule
Gemma-4-E4B & 53.2 & 38.3 & \cellcolor{up!20}+14.9 & 46.8 \\
Gemma-4-26B-A4B & 63.8 & 38.3 & \cellcolor{up!30}+25.5 & 50.0 \\
Gemma-4-31B & 74.5 & 36.7 & \cellcolor{up!43}+37.8 & 54.8 \\
MiMo v2.5 & 70.2 & 51.7 & \cellcolor{up!24}+18.5 & 59.5 \\
GPT-5.3-codex & 78.7 & 48.3 & \cellcolor{up!35}+30.4 & 61.9 \\
Gemini 3.1 Pro & 80.9 & 60.0 & \cellcolor{up!26}+20.9 & 71.4 \\
\bottomrule
\end{tabular}
\end{table}

We evaluate 21 open- and closed-weight VLM judges, including the Gemma-4-E4B base model of VisionQ-Judge.

\paragraph{The strongest judge reaches 68.6\%, above image-quality baselines.}
Gemini 3.1 Pro reaches the highest accuracy, 68.6\%, against a chance level of 32.2\%; within the Gemma-4 family, accuracy rises with size from 45\% (E4B) to 57\% (31B), and the weakest judge, Grok-4.20, reaches 39.8\% (\cref{tab:leaderboard}; all 21 judges in Tab.~S6, judge list in App.~\ref{sec:s-protocol}).
It leads each of the next four judges by 5.5 to 7.8 points (every CI above zero), whereas the benchmark does not separate these four (GPT-5.5 minus GPT-5.2: +1.9 points, CI $-$2.5 to +6.3).
A baseline that picks the crop with the highest no-reference quality score reaches at most 41.7\% (CLIP-IQA+~\cite{clipiqa}; MUSIQ~\cite{musiq} 40.1\%, TOPIQ~\cite{topiq} 40.8\%, Laplacian sharpness 36.2\%; App.~\ref{sec:s-results}), 27 points below the strongest judge.

\paragraph{Raters support the key, and judges agree with it more often where both raters pick it.}
On the 88 questions where both raters decide, they choose the key on 62.5\% and 68.2\%, against 32.7\% by chance.
These rates are close to how often the two raters choose the same crop (70.5\%, $\kappa=0.59$, \cref{tab:human}).
Of the 62 questions on which they agree, 47 are the key.
Both raters choose the same other crop on 15 of the 126 questions.
Agreement with the key also rises on the reference screen.
Every controlled judge agrees with the key more often when both raters pick it (47 questions) than when at least one picks another crop (60 questions).
Gemini 3.1 Pro, for example, scores 80.9\% against 60.0\%.
Even on these 47, the strongest judge disagrees with the key on about one question in five.

The controls (\cref{tab:controls}) compare conditions within one fresh run per judge (App.~\ref{sec:s-protocol}).

\paragraph{Choices change with the option order, most for the smallest judges (E1).}
Averaged over orders, accuracy moves by only $-$1.6 to +3.7 points, but judges pick the same crop in every order on 20.4\% (Gemma-4-E4B) to 64.7\% (Gemini 3.1 Pro) of questions, against 13.8\% at random.
A plurality vote over orders therefore falls below the standard score for five of the six judges, clearly so for the two smallest (Gemma-4-E4B: $-$8.7 points, CI $-$14.4 to $-$3.2; Gemma-4-26B-A4B: $-$6.1, CI $-$10.6 to $-$1.6).
The three Gemma judges give the last option close to the 30.0\% of answers expected from uniform guessing, so their order effect is instability rather than preference; only MiMo v2.5, GPT-5.3-codex and Gemini 3.1 Pro lean towards the last option (38.3\%, 38.6\% and 36.3\%).

\paragraph{Judges depend on the images but hardly on the named criterion (E2, E3).}
With blank tiles or another question's composite, every judge falls to between 26.9\% and 33.0\%, near chance; these are guesses rather than refusals, since only two judges ever decline to answer, on at most 8.7\% of questions.

Without an explicit criterion, judges lose just 1.6--8.7 points (resolved for three of six), retaining 86--97\% of their accuracy and remaining 7--32 points above chance. VisionQ-Judge shows the same pattern, and its attention changes little across criteria (\cref{sec:judge}; App.~\ref{sec:s-attention}).

\paragraph{Mirrored crops leave accuracy unchanged, which argues against recall of the published figures.}
Some question strings retain method names, and a judge might recall a published figure rather than compare its outputs.
Mirroring every crop keeps the relative quality of the outputs but breaks an exact match with the published figure.
This mirroring changes the accuracy of Gemini 3.1 Pro by $-$0.3 points (CI $-$3.5 to +2.8), of GPT-5.3-codex by +1.0 (CI $-$3.8 to +5.6) and of Gemma-4-31B by +1.9 (CI $-$3.4 to +7.1) (App.~\ref{sec:s-results}).

\begin{figure}[t]
\centering
\includegraphics[width=\columnwidth]{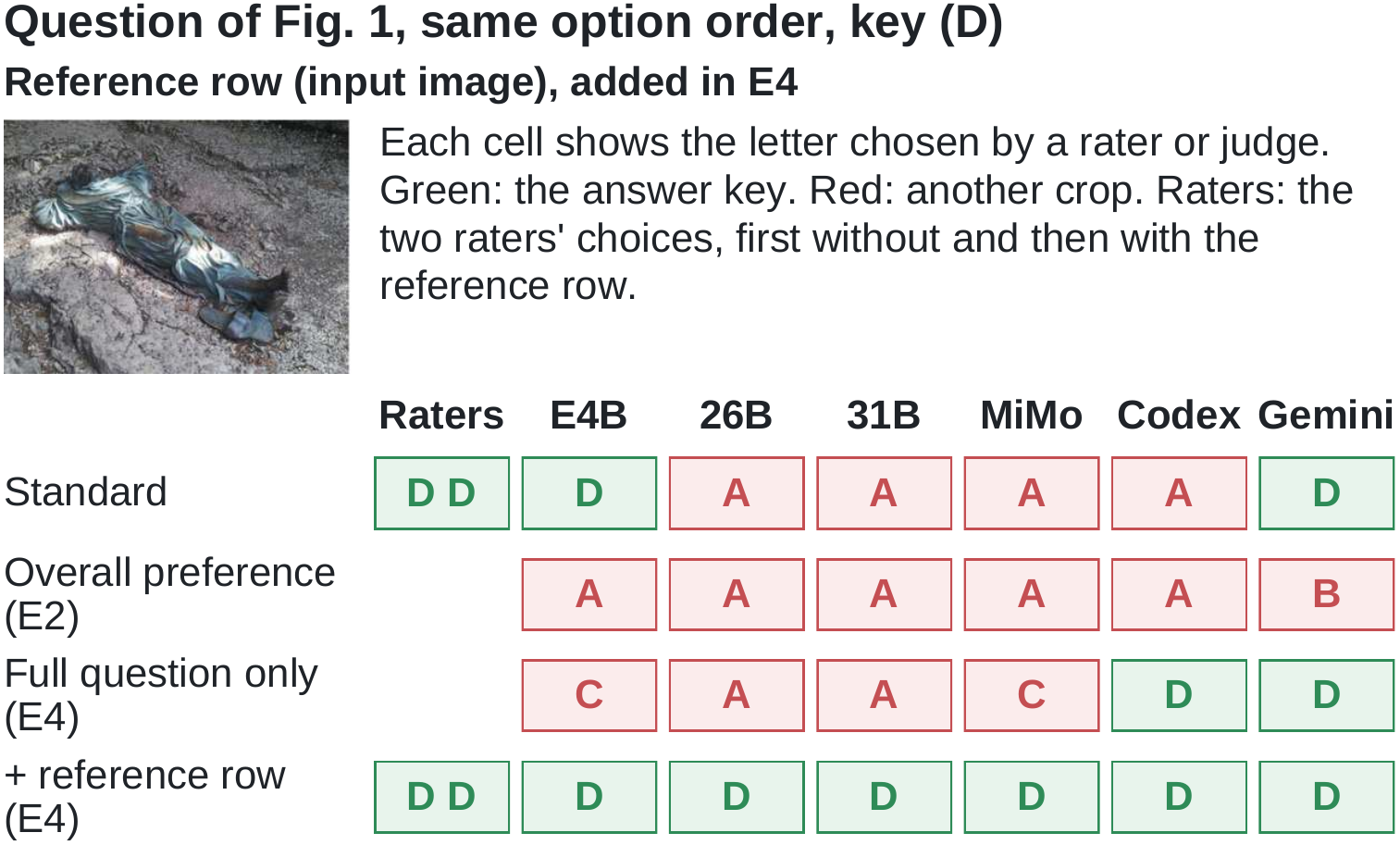}
\caption{Judge choices for the question of \cref{fig:overview}, with the same option order. The answer key, (D), is the source paper's method, and both raters choose it. Of the six judges tested under input controls, two choose (D) in the standard run and two with the full question alone. With the full question and the reference row, here the input image, all six choose (D) (E4). Columns: the two raters, Gemma-4-E4B, Gemma-4-26B-A4B, Gemma-4-31B, MiMo v2.5, GPT-5.3-codex and Gemini 3.1 Pro.}
\label{fig:teaser}
\end{figure}

\paragraph{Restoring the full question with the reference helps the larger judges and hurts the smallest (E4).}
With the full question alone, no judge changes by more than 2.6 points, and every interval includes zero.
On the 205 questions with a reference, adding the reference row lowers Gemma-4-E4B by 8.8 points (CI $-$16.2 to $-$1.0) but raises Gemma-4-26B-A4B, Gemma-4-31B and GPT-5.3-codex by 8.3 to 9.8 points (GPT-5.3-codex: CI +3.1 to +16.4), Gemini 3.1 Pro by 4.9 (CI +0.4 to +9.8) and MiMo v2.5 by 5.9.
\Cref{fig:teaser} shows this effect on one 3D reconstruction comparison: both raters pick the authors' winner, but only two of the six controlled judges do in the standard run; once the reference row is added, all six do.
The gain therefore comes from the reference image, which the standard protocol omits; even so, all six judges still score lower on Reference Fidelity than on Object Form (GPT-5.3-codex: 56\% against 72\%), as all 21 judges do under the standard condition.

\paragraph{Where every judge disagrees with the authors, the judges agree with one another.}
On 10 questions none of the 21 judges picks the key, and on 18 all of them do; yet on the 10, a median of 17 of them pick the same other crop, and of the six such questions in the human sample, at least one rater agrees with the judges on four.
Part of the judges' disagreement with the key is therefore systematic and shared by human readers.

\begin{figure*}[t]
\centering
\includegraphics[width=\textwidth]{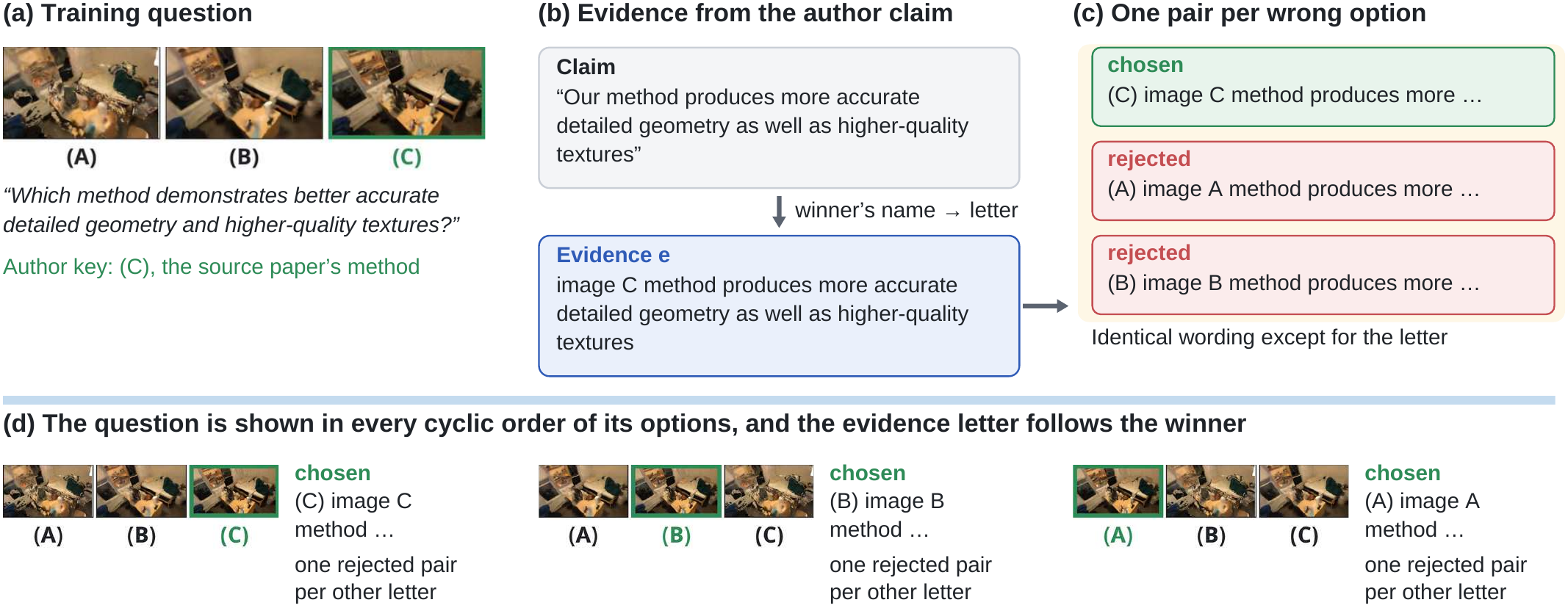}
\caption{From a training question to symmetric evidence pairs. (a)~Question with key (C). (b)~As in \cref{fig:overview}, the authors' claim identifies the winner. Here, the claim also supplies the evidence sentence $e$, with the winner's name replaced by its option letter. (c)~Chosen and rejected responses share $e$ and differ only in the letter. (d)~In each cyclic option order, the winner's letter changes to match its new position.}
\label{fig:pairs}
\end{figure*}
\begin{figure*}[t]
\centering
\includegraphics[width=\textwidth]{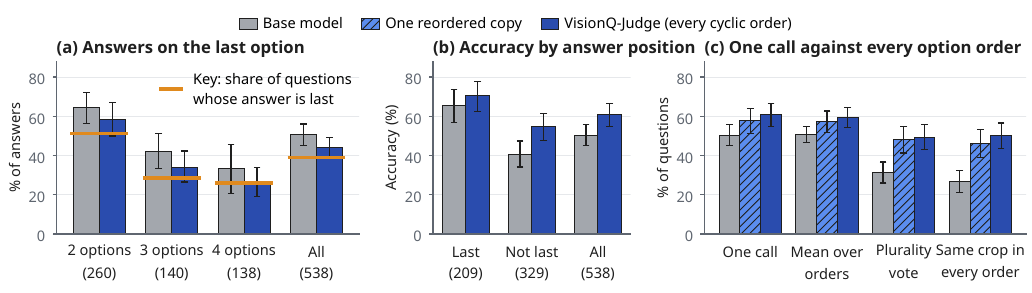}
\caption{VisionQ-Judge against its Gemma-4-E4B base model on the 538 VisionQ-MCQ test questions (mean of three seeds). Error bars: 95\% paper-bootstrap intervals of each bar; \cref{tab:judge} gives the paired changes. (a)~Share of answers on the last option; orange: questions whose answer is last. (b)~Accuracy by answer position. (c)~One call against answers pooled over every cyclic option order. \emph{One reordered copy}: each training question is reordered once instead of in every cyclic order.}
\label{fig:position}
\end{figure*}

\section{VisionQ-Judge}
\label{sec:judge}

On VisionQ-Bench, Gemma-4-E4B is unstable across option orders but shows no last-option preference (\cref{sec:results}).
VisionQ-MCQ prompts also include the figure caption and row name, and there the same model leans towards the last option: on the test split it gives the last option 50.7\% of its answers, against 38.8\% in the key, and it is right on 65.6\% of the questions whose answer is last but on 40.7\% of the others.
We therefore train and test VisionQ-Judge on VisionQ-MCQ, which is also large enough for a paper-disjoint split (App.~\ref{sec:s-judge}).

\paragraph{Data and splits.}
After capping each criterion leaf at 300 questions, VisionQ-MCQ is split once, before any training run, so that every paper lies on one side: 2{,}687 training, 359 validation and 538 test questions from 239, 79 and 168 papers, stratified by leaf so that each side's axis mix stays within 1.5 points of the pool's (App.~\ref{sec:s-judge}).
Every judge on this split receives the same prompt (App.~\ref{sec:s-judge}).

\paragraph{Symmetric evidence pairs.}
Each pair shares one evidence sentence from the author claim, and the preferred and rejected responses differ only in the letter it refers to: $y_{\text{chosen}} = (w)\,e_w$ and $y_{\text{rejected}} = (r)\,e_r$, with one pair for every wrong option $r$ (\cref{fig:pairs}); where the evidence names no letter, both responses carry it unchanged.
The rejected evidence replaces each mention of the winner's letter with the loser's, so the two responses share their wording apart from one letter and the objective gives no reward for longer answers.
Each training question is rendered in every cyclic order of its options, so the winner occupies each position equally often (13{,}394 pairs), and we fine-tune Gemma-4-E4B-it with LoRA~\cite{lora} on its language and vision towers for one epoch of DPO; the pairs are built without any language-model calls (App.~\ref{sec:s-judge}).

\begin{table}[t]
\centering
\caption{VisionQ-Judge on the 538 VisionQ-MCQ test questions. $\Delta$: paired accuracy difference from the base model in the same block; in block (b), VisionQ-Judge minus the listed judge; * 95\% paper-bootstrap interval excludes zero. \emph{Last}: percentage of answers on the last option; \colorbox{down!35}{red} shading grows with the excess over 38.8\%, the share of questions whose key is last. \emph{Calls}: model calls per question. Trained models: mean of three seeds. Blocks (a) and (c) ran on different GPUs (App.~\ref{sec:s-judge}); compare within a block.}
\label{tab:judge}
\footnotesize
\setlength{\tabcolsep}{1.7pt}
\begin{tabular}{@{}lcccc@{}}
\toprule
\textbf{Judge} & \textbf{Calls} & \textbf{Acc.} & $\boldsymbol{\Delta}$ & \textbf{Last} \\
\midrule
\multicolumn{5}{@{}l}{\emph{(a) Fine-tuning Gemma-4-E4B}} \\
Base model & 1 & 50.4 & -- & \cellcolor{down!29}50.7 \\
No reordering (seed 42) & 1 & 52.4 & +2.0 {\scriptsize[$-$2.5,+6.5]} & \cellcolor{down!27}50.0 \\
One reordered copy & 1 & 57.7 & \cellcolor{up!23}+7.3* {\scriptsize[+1.9,+12.4]} & \cellcolor{down!22}47.5 \\
\rowcolor{bench!12}VisionQ-Judge (all orders) & 1 & \textbf{61.0} & \cellcolor{up!32}\textbf{+10.6*} {\scriptsize[+5.0,+16.0]} & \cellcolor{down!15}43.9 \\
\midrule
\multicolumn{5}{@{}l}{\emph{(b) Larger judges, same prompt;} $\Delta$ = VisionQ-Judge minus the judge} \\
MiMo v2.5 & 1 & 55.0 & \cellcolor{up!20}+5.9* {\scriptsize[+1.2,+11.0]} & 34.0 \\
Gemma-4-26B-A4B & 1 & 56.5 & +4.5 {\scriptsize[$-$0.7,+10.0]} & \cellcolor{down!43}58.0 \\
Gemma-4-31B & 1 & 63.4 & $-$2.4 {\scriptsize[$-$7.5,+2.9]} & \cellcolor{down!31}51.9 \\
GPT-5.5 & 1 & 64.9 & $-$3.9 {\scriptsize[$-$9.1,+1.1]} & \cellcolor{down!35}53.9 \\
GPT-5.3-codex & 1 & 65.1 & $-$4.1 {\scriptsize[$-$9.2,+0.8]} & \cellcolor{down!37}54.6 \\
\midrule
\multicolumn{5}{@{}l}{\emph{(c) Inference-time alternatives (other GPU type)}} \\
Base model & 1 & 47.0 & -- & \cellcolor{down!13}42.9 \\
\quad + PriDe~\cite{wang2023large} & 1 & 48.0 & +0.9 {\scriptsize[$-$0.4,+2.4]} & \cellcolor{down!14}43.5 \\
\quad + log-prob.\ averaging & 2.8 & 56.7 & \cellcolor{up!29}+9.7* {\scriptsize[+3.9,+15.4]} & 37.2 \\
\rowcolor{bench!12}VisionQ-Judge & 1 & 57.4 & \cellcolor{up!31}+10.3* {\scriptsize[+4.8,+15.9]} & \cellcolor{down!11}41.8 \\
\rowcolor{bench!12}\quad + log-prob.\ averaging & 2.8 & \textbf{61.2} & \cellcolor{up!40}\textbf{+14.2*} {\scriptsize[+8.0,+20.1]} & 36.4 \\
\bottomrule
\end{tabular}
\end{table}

\paragraph{Results.}
Over three seeds, VisionQ-Judge raises test accuracy from 50.4\% to 61.0\% (\cref{tab:judge}a), and the validation split agrees (+10.2 points, CI +4.6 to +16.7).
It lowers the last-option share from 50.7\% to 43.9\% (\cref{fig:position}a; $-$6.9 points, CI $-$11.9 to $-$1.5), whereas Gemma-4-26B-A4B, Gemma-4-31B and both GPT judges give the last option 52 to 58\% of their answers (\cref{tab:judge}b).
Most of the gain comes from the questions whose answer is not last: on these 329, accuracy rises from 40.7\% to 54.9\% (+14.2 points, CI +5.5 to +22.0), against +4.9 (CI $-$1.2 to +11.7) on the 209 whose answer is last (\cref{fig:position}b).
The tuned 4B model scores above MiMo v2.5 (+5.9 points, CI +1.2 to +11.0) and Gemma-4-26B-A4B (+4.5, CI $-$0.7 to +10.0) and cannot be separated from the other three (largest gap, to GPT-5.3-codex: $-$4.1, CI $-$9.2 to +0.8; \cref{tab:judge}b).
As on VisionQ-Bench, naming the criterion adds little: when asked which output is better overall, neither the base model nor VisionQ-Judge loses accuracy (App.~\ref{sec:s-judge}).

\paragraph{Training, averaging and reordering.}
Log-probability averaging over every cyclic order raises the base model by 9.7 points, to the one-call level of VisionQ-Judge, but at 2.8 calls per question (\cref{tab:judge}c; App.~\ref{sec:s-judge}).
PriDe~\cite{wang2023large}, a single-call correction of the position prior, adds 0.9 points, and VisionQ-Judge stays 9.4 points ahead of it (CI +3.8 to +15.2).
The position prior that PriDe estimates is close to uniform, which leaves little to correct (App.~\ref{sec:s-judge}).
The reordering drives the gain: without reordered copies the model gains 2.0 points and keeps the base model's last-option share, one reordered copy gives +7.3, and every cyclic order gives +10.6 (+3.3 over one copy, CI +1.0 to +5.8; \cref{fig:position}c), although these regimes also differ in the number of training pairs.

\section{Conclusion}
\label{sec:conclusion}

VisionQ asks whether VLM judges can take over the comparison of computer vision methods from human readers.
The strongest judge reaches an expert alignment of 68.6\%, against 32.2\% by chance. Judges agree with the key more often where blinded raters do, and they depend on the images.
They gain little from the named criterion, however, and even the strongest misses one question in five where both raters agree with the authors.
Three implications follow for a judge that compares method outputs.
Its per-criterion scores should not be read as evidence about that criterion alone, since an overall-preference question keeps most of its agreement with the authors.
It should see the reference image when the criterion concerns fidelity to that image, since restoring the reference raises four of the five larger judges by 4.9 to 9.8 points.
A small judge should be checked across option orders or trained to be stable, since its choices change with the order.
VisionQ-Judge, a 4B judge trained on every cyclic order of the options, reaches 61.0\% on the VisionQ-MCQ test split, above the 56.5\% of a 26B model from the same family.
With the questions, controls and weights released, new VLM judges can be tested on the same evidence.
The next step is to teach judges to act on the named criterion, towards objective, criterion-specific comparisons of past, present and future methods. Creation of tools and skills based on VisionQ will be useful for the Computer Vision community in research, comparison and writing.

\paragraph{Limitations.}
VisionQ measures agreement with the claims that authors publish.
This design makes every answer traceable to a statement in the source paper, but it also ties the key to the source paper: the claimed winner is likely the authors' method, and no question has a baseline winner.
VisionQ therefore does not yet test whether a judge's answer follows a change of criterion.
A criterion-switch set, in which two criteria favour different outputs of the same row, would measure this directly.
The questions come from static figures in CVPR and ICCV 2023--2024, and VisionQ-Judge starts from one base model, so video, interactive 3D results and other base models remain to be covered.
The axes are uneven: of the 309 questions, Prompt Match has 22 and Scene Layout 8, so the scores on these two axes rest on few questions.
App.~\ref{sec:s-human} covers the licensing of the release and rater consent.

\section*{Acknowledgements}
We thank the authors of the source papers, whose qualitative comparisons and published claims are the basis of VisionQ.
This publication has emanated from research conducted with the financial support of Taighde \'Eireann -- Research Ireland under Grant number 23/RC/13506. For the purpose of open access, the author has applied a CC BY public copyright licence to any Author Accepted Manuscript version arising from this submission.

\clearpage
\appendix
\renewcommand{\thetable}{S\arabic{table}}\setcounter{table}{0}
\renewcommand{\thefigure}{S\arabic{figure}}\setcounter{figure}{0}
\newcommand{\main}[1]{Sec.~#1}

The appendices give the material that the main text refers to: the full criterion vocabulary (\cref{sec:s-vocab}), how the scored items were selected and audited (\cref{sec:s-items}), the evaluation protocol and judges (\cref{sec:s-protocol}), additional results (\cref{sec:s-results}), the human study (\cref{sec:s-human}), VisionQ-Judge (\cref{sec:s-judge}) and the judges' criterion sensitivity, position bias and visual grounding (\cref{sec:s-attention}). Tables and figures numbered with the prefix ``S'' are in the appendices.

\section{Criterion vocabulary}
\label{sec:s-vocab}

\paragraph{Construction.}
Open coding of 9{,}228 claims from 1{,}409 CVPR and ICCV papers gave each claim a short label naming the criterion it judges; merging duplicate and synonymous labels left 447 candidate criteria.
Following Clio~\cite{tamkin2024clio}, the labels were embedded and clustered into 60 leaf candidates, an LLM named each cluster by contrast with its neighbours, and the same procedure grouped the leaf candidates into 16 mid-level groups and these into six axes.
Three LLMs (GPT-5.5 Thinking, Gemini 3.1 Pro and Claude Opus 4.7) drafted the codes and cluster names independently, and human reviewers compared the drafts and wrote every final label.
Reviewers then checked each leaf against example claims, splitting or merging overlapping leaves and removing sparse ones, which left 51 leaves.
The mid-level groups were used only during construction.

\paragraph{One axis per claim.}
When a claim plausibly fits several axes, it receives exactly one by a fixed priority order: Prompt Match, Reference Fidelity, Relation, Scene Layout, Object Form, Image Appearance.
The order puts explicit conditioning signals first (a text prompt, then a visible reference) and then proceeds from relational and scene-level criteria to object- and image-level ones.
For example, a relation required by the prompt is assigned to Prompt Match, while a relation between visible entities without a prompt is assigned to Relation; a shape judged against a visible target is assigned to Reference Fidelity, while its quality without a reference is assigned to Object Form.
Two rules close known boundary cases: a claim goes to Relation as a part-whole relation only when two visible parts are judged jointly, and to Reference Fidelity only when the reference is visible and named in the figure.

\paragraph{Out-of-scope claims.}
Claims that cannot be checked in the figure receive one of nine reason codes and are excluded from the vocabulary and the benchmark: temporal (motion or video), non-visual method property, metric calibration, latent or internal representation, claims that require multiple samples, dataset or training, protocol robustness, claims that require external knowledge, and claims not visible in the figure.
Of 1{,}486 annotated figures, 1{,}426 (96.0\%) fall inside the vocabulary.
Fields lean on different leaves: papers on 3D reconstruction place 57.8\% of their claims on three Object Form leaves, whereas papers on 2D generation place 30.2\% on one Prompt Match leaf.

\paragraph{Axes.}
\textbf{Image Appearance}: low-level appearance of a whole panel or crop (blur, noise, exposure, contrast, color, lighting, global artifacts, realism, style); excludes object shape, spatial relations, prompt and reference matching.
\textbf{Object Form}: internal quality of visible objects, regions, bodies, faces, surfaces and masks; excludes scene arrangement, pairwise relations and matching to a prompt or reference.
\textbf{Scene Layout}: scene-level spatial organisation and physical arrangement; excludes object-internal defects and pairwise relation claims.
\textbf{Relation}: relationships between visible entities, parts, attributes or correspondence points; a single malformed part of one entity belongs to Object Form.
\textbf{Reference Fidelity}: whether an output preserves or matches a visible reference (source image, target, ground truth, annotation, mask, landmark or before/after state); a comparison between methods without a visible reference is not Reference Fidelity.
\textbf{Prompt Match}: agreement with a text prompt, caption or instruction (object presence, attributes, spatial instructions, semantics, text, counts, actions, expressions).
\Cref{tab:s-leaves1,tab:s-leaves2} list the 51 leaves with their working definitions.

\begin{table*}[t]
\centering
\caption{Criterion leaves (part 1 of 2): Image Appearance, Object Form and Scene Layout.}
\label{tab:s-leaves1}
\footnotesize
\setlength{\tabcolsep}{3pt}
\begin{tabular}{@{}p{2.3cm}p{0.2cm}p{2.9cm}p{11.4cm}@{}}
\toprule
\textbf{Axis} & & \textbf{Leaf} & \textbf{Working definition} \\
\midrule
Image Appearance &  & Blur & Focus loss, smear, or blur visible at the panel or crop level. \\
 &  & Noise & Random speckle, grain, sensor-like corruption, or noisy visual artifacts. \\
 &  & Exposure & Over-exposure, under-exposure, saturation clipping, or visibility loss from brightness. \\
 &  & Contrast & Insufficient or excessive tonal separation that harms visible interpretation. \\
 &  & Color & Color cast, color inconsistency, or visually implausible color independent of prompt/reference criteria. \\
 &  & Lighting & Illumination, shadow, reflection, or lighting quality visible in the output. \\
 &  & Global Artifact & Panel-wide artifact not better explained by object form, layout, relation, prompt, or reference match. \\
 &  & Realism & Overall perceptual plausibility or photorealism when not tied to a specific object, prompt, or reference. \\
 &  & Style & Global visual style or aesthetic quality when judged directly rather than as reference or prompt match. \\
\midrule
Object Form &  & Shape & Object geometry, silhouette, deformation, or structural shape correctness. \\
 &  & Boundary & Object or region edge quality, contour sharpness, silhouette completeness, or boundary artifacts. \\
 &  & Texture & Local texture quality, texture continuity, or texture corruption on visible regions. \\
 &  & Surface & Surface smoothness, topology, wrinkles, holes, or visible surface defects. \\
 &  & Material & Material appearance or material property quality of an object or region. \\
 &  & Completeness & Missing, truncated, duplicated, or incomplete object parts. \\
 &  & Pose & Pose plausibility of a single visible object, body, or articulated entity when not judged against ground truth. \\
 &  & Face & Face-specific form quality including expression quality when not prompt- or reference-driven. \\
 &  & Anatomy & Body, limb, hand, joint, or anatomical plausibility of one entity. \\
 &  & Detail & Fine detail, thin structures, small parts, or local feature preservation. \\
 &  & Segmentation & Mask or segmentation visual quality when judged as object-region form rather than ground-truth match. \\
\midrule
Scene Layout &  & Composition & Overall arrangement of regions, objects, foreground, background, or panel contents. \\
 &  & Depth & Depth ordering, 3D arrangement, or perceived scene depth. \\
 &  & Scale & Relative or absolute size plausibility at scene level. \\
 &  & View & Viewpoint, novel-view layout, or multi-view scene-level consistency. \\
 &  & Spatial Distribution & Distribution, spacing, clumping, or spread of visible scene elements or point samples. \\
 &  & Occlusion Layout & Occlusion ordering or impossible scene-level overlap patterns. \\
 &  & Camera Geometry & Camera perspective, projection geometry, or viewpoint geometry visible in the figure. \\
 &  & Physical Plausibility & Scene-level physical plausibility such as floating objects, floor contact at scene scale, or impossible arrangements. \\
\bottomrule
\end{tabular}
\end{table*}

\begin{table*}[t]
\centering
\caption{Criterion leaves (part 2 of 2): Relation, Reference Fidelity and Prompt Match. For Reference Fidelity, the second column gives the internal split into source preservation (S) and target fidelity (T).}
\label{tab:s-leaves2}
\footnotesize
\setlength{\tabcolsep}{3pt}
\begin{tabular}{@{}p{2.3cm}p{0.2cm}p{2.9cm}p{11.4cm}@{}}
\toprule
\textbf{Axis} & & \textbf{Leaf} & \textbf{Working definition} \\
\midrule
Relation &  & Contact & Whether two visible entities or parts touch, connect, intersect, or maintain contact correctly. \\
 &  & Correspondence & Visible correspondence between entities, parts, points, views, or panels. \\
 &  & Attribute Binding & Correct binding of visible attributes to the right entity or part. \\
 &  & Spatial Relation & Pairwise relation such as left/right, above/below, inside/outside, near/far, or in-front/behind. \\
 &  & Interaction & Visible interaction between people, objects, agents, tools, or scene elements. \\
 &  & Part-Whole Relation & Relation between two or more visible parts judged jointly (e.g., wheel misaligned with axle). \\
\midrule
Reference Fidelity & S & Identity Preservation & Preservation of subject/person/object identity against a visible source/reference image. \\
 & S & Background Preservation & Preservation of background content from a visible source/reference image. \\
 & S & Edit Preservation & Preservation of non-edited content in before/after or edit-reference setups. \\
 & S & Style Fidelity & Style copied from a visible reference image (DreamBench-style transfer). NOT for prompt-requested style or intrinsic rendering style. \\
 & T & Reconstruction Fidelity & Reconstruction fidelity against a target, source, or ground truth (e.g., PSNR/SSIM/LPIPS-style judgments on visible reference). \\
 & T & Geometry Fidelity & Geometric fidelity to a visible source, target, annotation, or ground truth. \\
 & T & Mask Fidelity & Mask fidelity against a visible mask, annotation, or ground truth. \\
 & T & Landmark Fidelity & Keypoint, landmark, or pose-coordinate fidelity against visible ground truth or annotation. \\
 & T & Segmentation Fidelity & Segmentation output fidelity against visible ground-truth segmentation. \\
\midrule
Prompt Match &  & Object Presence & Requested object, entity, category, or scene element is present or absent as specified. \\
 &  & Attribute Match & Requested color, material, style, state, or other attribute is correctly assigned. \\
 &  & Spatial Instruction & Prompt-specified spatial arrangement or relation is satisfied. \\
 &  & Semantic Match & Overall semantic content matches the caption, prompt, or instruction. \\
 &  & Text Legibility & Requested text, glyphs, symbols, or readable writing are rendered correctly. \\
 &  & Count Match & Requested number of objects or entities is correct. \\
 &  & Action Match & Requested action, activity, or event is depicted correctly. \\
 &  & Expression Match & Requested facial expression, emotion, or pose state is depicted correctly. \\
\bottomrule
\end{tabular}
\end{table*}

\section{Item selection and audit}\label{sec:s-items}

\begin{figure*}[t]
\centering
\includegraphics[width=\textwidth]{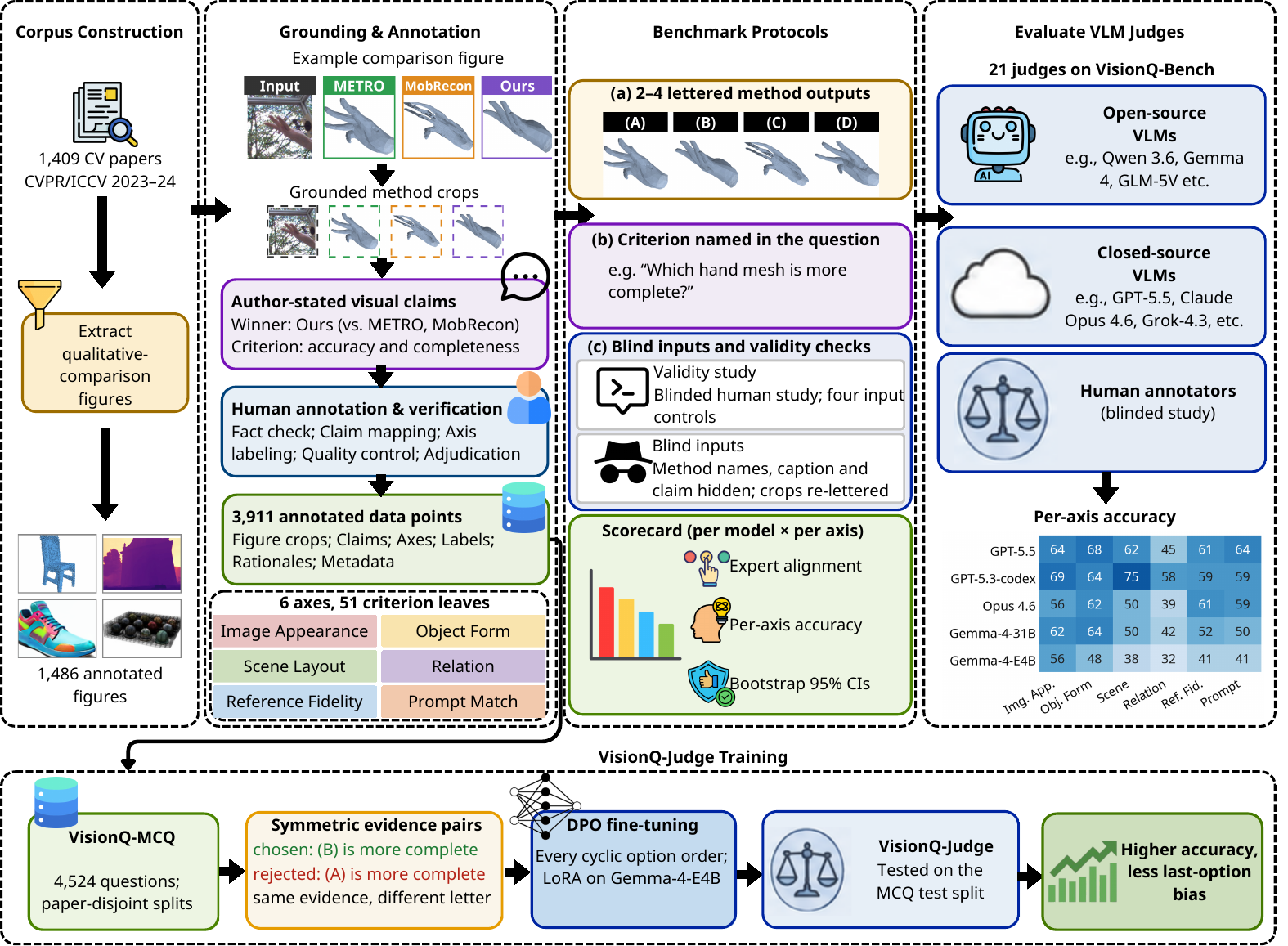}
\caption{Full VisionQ pipeline (complement to Fig.~2). Author claims in comparison figures become blind questions that name one criterion (\main{3}). Twenty-one judges answer these questions, and a human study and four input controls test what their accuracy rests on (\main{4}). VisionQ-MCQ, built from the same annotations, trains VisionQ-Judge (\main{6}).}
\label{fig:s-workflow}
\end{figure*}

\begin{table}[t]
\centering
\caption{From the source corpus to the scored questions (\main{3.2}). A data point is one comparison row linked to one author claim. VisionQ-MCQ splits: after capping each leaf at 300 questions.}
\label{tab:items}
\scriptsize
\setlength{\tabcolsep}{2.5pt}
\begin{tabular}{@{}lrrrr@{}}
\toprule
\textbf{Population} & \textbf{Papers} & \textbf{Figures} & \textbf{Data points} & \textbf{Questions} \\
\midrule
Source corpus & 1{,}409 & & & \\
Annotated pool & 810 & 1{,}486 & 3{,}911 & \\
\addlinespace[2pt]
\multicolumn{5}{@{}l}{\emph{VisionQ-Bench}} \\
\quad All questions & 114 & 132 & & 332 \\
\quad Rule-filtered, scored & 109 & & & 309 \\
\addlinespace[2pt]
\multicolumn{5}{@{}l}{\emph{VisionQ-MCQ}} \\
\quad All questions & 513 & 707 & 1{,}737 & 4{,}524 \\
\quad Training split & 239 & & & 2{,}687 \\
\quad Validation split & 79 & & & 359 \\
\quad Test split & 168 & & & 538 \\
\bottomrule
\end{tabular}
\end{table}
\Cref{fig:s-workflow} shows the full pipeline, and \cref{tab:items} traces the corpus to the scored questions and the VisionQ-MCQ splits.

\paragraph{Rule-based filter.}
VisionQ-Bench draws 332 questions from 114 papers.
A filter on option labels removes 23: 6 in which two options carry the same method name, 9 in which an option is labelled as a ground-truth, input or other context panel, and 8 left with fewer than two method families once such panels are set aside.
This leaves the 309 scored questions from 109 papers.

\paragraph{Option-label audit.}
Every scored question was then read against its option labels, the published question, the caption and the author claim, and checked for six possible issues (\cref{tab:s-audit}).
F1 and F2 can make the key harder to define: a context panel among the options can be picked by a judge, and options of different visualisation types (\eg a depth map next to an RGB image) are hard to compare on one criterion.
The 59 questions with F1 or F2 were checked by hand against their source figures; the flag was upheld for 55 and lifted for 4.
The audited subset removes the 55, leaving 254 questions from 90 papers.
This subset serves as a sensitivity check.
The main text reports all results on the 309 questions, and its conclusions also hold on the subset (\cref{tab:leaderboard-full,tab:controls-audited}).
F3 to F6 leave the key well defined and are kept.
Control E4 addresses F3, and F4 where a reference image exists, since it removes method names from the question and adds the reference row; F5 and F6 do not change which output the authors claim is best.

\begin{table}[t]
\centering
\caption{Flags of the option-label audit on the 309 scored questions. A question can carry several flags.}
\label{tab:s-audit}
\footnotesize
\setlength{\tabcolsep}{3pt}
\begin{tabular}{@{}llr@{}}
\toprule
& \textbf{Flag} & \textbf{Questions} \\
\midrule
F1 & a reference or input panel among the options & 59 \\
F2 & options of different visualisation types & 7 \\
F3 & the question names a method & 18 \\
F4 & the answer needs a reference or prompt & 40 \\
F5 & several options are own-method variants & 33 \\
F6 & other issue noted on reading & 13 \\
\midrule
\multicolumn{2}{@{}l}{F1 or F2 (checked by hand; 55 confirmed)} & 59 \\
\bottomrule
\end{tabular}
\end{table}

\paragraph{Question text.}
The standard protocol uses the published question string, which is cut at a fixed length for 126 of the 309 questions.
For E4 and the human study, the full question is restored from the source annotation and sanitised: method names and comparison targets (\eg ``compared to MobRecon'') are removed so that the question names only the criterion.

\section{Evaluation protocol}
\label{sec:s-protocol}

\paragraph{Rendering.}
The candidate crops of a question are rendered into one composite image.
Each crop is resized to at most 512\,px on its longer side and placed under a 64\,px banner with its letter; two or three crops are placed in one row and four in a $2{\times}2$ grid.
In E4, the reference crops of the source row are rendered in a separate row above the candidates under grey banners with their labels (\eg GT, Input).
Composites are downscaled to at most 1{,}024\,px on the longer side before they are sent.

\paragraph{Prompt.}
Every judge receives the same system prompt and one user message with the composite, at temperature 0:
\begin{quote}\footnotesize\ttfamily\raggedright
System: You are a careful image-quality judge. You will see a question and between 2 and 4 candidate images, each labeled (A), (B), (C), (D). Pick the SINGLE image that best satisfies the question. Reply with EXACTLY one of: (A), (B), (C), (D). Do not explain. Do not add any other text.\\[3pt]
User: Question: \{question\}\\
The single image below contains \{n\} candidate tiles arranged in a grid. Each tile has a label banner above it: (A), (B), \ldots{} Reply with EXACTLY one of those letters in parentheses.
\end{quote}
With reference tiles (E4), the user message states that the image has two sections, that the top row holds reference imagery for context only and must not be picked, and that the judge picks one of the candidate tiles below.

\paragraph{Parsing.}
The answer is the last parenthesised letter within the range of options in the response, which handles reasoning models that end with their answer; if there is none, the last bare letter is used.
Responses without a valid letter, and failed requests after retries, count as wrong.

\paragraph{Judges.}
\Cref{tab:s-judges} lists the judges and the identifiers through which they were queried.
The controls of \main{4} are fresh runs with the same identifiers.
Gemini 3.1 Pro was queried after the other judges, through its preview identifier; in the controls and the mirror test its output is capped at 8{,}192 tokens, which no response reaches.
The runner sets temperature 0 and passes no reasoning options, so each judge reasons as its provider's default dictates: GPT-5.4, GPT-5.1, the Claude models, Grok-4.20, Mistral Medium 3.5, Gemma-4-31B and Gemma-4-26B-A4B answer with a few output tokens, while the other judges produce a reasoning trace before the answer.

\begin{table*}[t]
\centering
\caption{Judges and model identifiers. Claude models were queried through Amazon Bedrock and the other API judges through OpenRouter.}
\label{tab:s-judges}
\footnotesize
\setlength{\tabcolsep}{4pt}
\def\UrlBreaks{\do\/\do\-}\urlstyle{tt}
\begin{tabularx}{\textwidth}{@{}l>{\raggedright\arraybackslash}X@{\hspace{14pt}}l>{\raggedright\arraybackslash}X@{}}
\toprule
\textbf{Judge} & \textbf{Identifier} & \textbf{Judge} & \textbf{Identifier} \\
\midrule
GPT-5.5/5.4/5.3-codex/5.2/5.1 & \texttt{openai/gpt-5.\{5,4,3-codex,2,1\}} & Qwen 3.6-35B-A3B & \nolinkurl{qwen/qwen3.6-35b-a3b} \\
Claude Opus 4.6 & \nolinkurl{anthropic.claude-opus-4-6-v1} & Qwen 3.6-27B & \nolinkurl{qwen/qwen3.6-27b} \\
Claude Sonnet 4.6 & \nolinkurl{anthropic.claude-sonnet-4-6} & Qwen 3.5-9B & \nolinkurl{qwen/qwen3.5-9b} \\
Claude Haiku 4.5 & \nolinkurl{anthropic.claude-haiku-4-5-20251001-v1:0} & Nemotron-Omni & \nolinkurl{nvidia/nemotron-3-nano-omni-30b-a3b-reasoning} \\
Grok-4.3 / Grok-4.20 & \nolinkurl{x-ai/grok-4.3}, \nolinkurl{x-ai/grok-4.20} & Gemma-4-31B & \nolinkurl{google/gemma-4-31b-it} \\
MiMo v2.5 & \nolinkurl{xiaomi/mimo-v2.5} & Gemma-4-26B-A4B & \nolinkurl{google/gemma-4-26b-a4b-it} \\
GLM-5V Turbo & \nolinkurl{z-ai/glm-5v-turbo} & Gemma-4-E4B & \nolinkurl{google/gemma-4-E4B-it} \\
Mistral Medium 3.5 & \nolinkurl{mistralai/mistral-medium-3-5} & Gemini 3.1 Pro & \nolinkurl{google/gemini-3.1-pro-preview} \\
\bottomrule
\end{tabularx}
\end{table*}

\paragraph{Statistics.}
All intervals are 95\% percentile intervals from a bootstrap over source papers: each resample draws papers with replacement and keeps all of their questions.
Leaderboard intervals use 4{,}000 resamples and all other intervals 10{,}000.
Paired differences (between conditions or judges) are computed per question and resampled the same way, so that questions from one paper stay together.

\paragraph{Input controls.}
Each control keeps the benchmark's renderer, prompt and parser and changes the input in one respect (E4 in two, which a text-only variant separates).
E1 renders every cyclic order of the options (1{,}029 composites).
A question with options (A, B, C) is also asked with the crops in the orders (B, C, A) and (C, A, B), so that the key's crop appears once in every position; the first order is the standard run.
Each answer is mapped back to the crop that its letter names, and E1 reports four scores.
\emph{Mean over orders} is the share of orders in which the judge picks the key's crop, averaged over questions.
A \emph{plurality vote} takes the crop picked in the most orders as the judge's answer; the question counts as correct only if this crop is the key's and no other crop has as many votes, so ties and unparseable answers count as wrong.
\emph{Same crop} is the share of questions on which the judge picks one crop in every order, whether or not it is the key's; a judge that answers at random would reach 13.8\% on these questions.
The last score is the share of answers on the last option, which is compared with always-first and always-last baselines.

For example, the question in Fig.~1 has four options, and the key, the paper's own method, is at (D) in the standard run.
Gemini 3.1 Pro picks the key in all four orders, so its mean over orders is 1, its plurality vote is correct, and it picks the same crop in every order.
MiMo v2.5 picks PatchmatchNet in two orders and ACMMP and the key in one each, so its mean over orders is 1/4 and its plurality vote selects PatchmatchNet and counts as wrong.

E2 replaces the question with ``Which output is better overall?'', so its accuracy estimates what a general preference would score against the same key.
For the question in Fig.~1, this replaces ``Which method demonstrates better geometry estimations and reconstruction accuracy'' with the overall question.
E3 asks the original question about white tiles and about a composite from another question with the same number of options.
On blank tiles, only Gemma-4-26B-A4B and Gemma-4-31B decline to answer, on 8.7\% and 7.1\% of questions, and these answers count as wrong.
E4 changes two parts of the input, and a text-only variant separates them.
First, it restores the full question, which the standard protocol cuts at a fixed length, and then removes method names and comparison targets from it.
In Fig.~1, the standard question ends with ``\ldots compared to COLMAP, N\ldots?'', and the restored question ends with ``\ldots reconstruction accuracy?''.
Second, it adds the reference row of the source figure above the candidates, here the input image, where such a row exists and could be cropped (205 of the 211 questions with a reference).
The six controlled judges span the leaderboard: Gemma-4-E4B (base), Gemma-4-26B-A4B (48.5\%), Gemma-4-31B (57.3\%), MiMo v2.5 (59.2\%, a reasoning model from another developer), GPT-5.3-codex (63.1\%) and Gemini 3.1 Pro (68.6\%). GPT-5.5 ties with GPT-5.3-codex on the benchmark and on VisionQ-MCQ (64.9\% against 65.1\%, difference $-$0.2 points, CI $-$3.3 to +2.9), so we run the controls for one of the two.
The fresh standard runs of the controls lie within 5.2 points of the leaderboard.
\section{Additional results}
\label{sec:s-results}

\paragraph{Per-axis scores.}
\Cref{tab:leaderboard-full} gives every judge's per-axis score with its interval; Tab.~2 shows nine judges and point estimates only.

\begin{table*}[t]
\centering
\caption{Expert alignment (\%) of every judge on VisionQ-Bench, overall, on the audited subset and per axis, with 95\% paper-bootstrap intervals (complement to Tab.~2). \emph{Audited}: the 254 scored questions that pass the human check of \cref{sec:s-items}. The second header row gives the number of questions / source papers.}
\label{tab:leaderboard-full}
\scriptsize
\setlength{\tabcolsep}{2.4pt}
\begin{tabularx}{\textwidth}{@{}l*{8}{>{\centering\arraybackslash}X}@{}}
\toprule
\textbf{Judge} & \textbf{Overall} & \textbf{Audited} & \textbf{Object Form} & \textbf{Ref.\ Fidelity} & \textbf{Img.\ Appearance} & \textbf{Relation} & \textbf{Prompt Match} & \textbf{Scene Layout} \\
 & 309 / 109 & 254 / 90 & 143 / 50 & 66 / 23 & 39 / 16 & 31 / 8 & 22 / 9 & 8 / 4 \\
\midrule
Gemini 3.1 Pro & 69 {\scriptsize[62,74]} & 72 {\scriptsize[66,79]} & 76 {\scriptsize[66,84]} & 59 {\scriptsize[49,69]} & 69 {\scriptsize[50,87]} & 52 {\scriptsize[43,64]} & 77 {\scriptsize[53,100]} & 62 {\scriptsize[29,100]} \\
GPT-5.3-codex & 63 {\scriptsize[57,69]} & 66 {\scriptsize[59,72]} & 64 {\scriptsize[55,73]} & 59 {\scriptsize[45,71]} & 69 {\scriptsize[53,83]} & 58 {\scriptsize[46,78]} & 59 {\scriptsize[29,88]} & 75 {\scriptsize[40,100]} \\
GPT-5.5 & 63 {\scriptsize[57,69]} & 67 {\scriptsize[61,73]} & 68 {\scriptsize[58,77]} & 61 {\scriptsize[48,72]} & 64 {\scriptsize[48,79]} & 45 {\scriptsize[31,60]} & 64 {\scriptsize[38,85]} & 62 {\scriptsize[29,100]} \\
GPT-5.2 & 61 {\scriptsize[55,67]} & 62 {\scriptsize[56,69]} & 66 {\scriptsize[57,75]} & 55 {\scriptsize[43,66]} & 62 {\scriptsize[46,76]} & 58 {\scriptsize[48,75]} & 59 {\scriptsize[29,84]} & 50 {\scriptsize[20,100]} \\
GPT-5.4 & 61 {\scriptsize[55,67]} & 63 {\scriptsize[57,70]} & 64 {\scriptsize[56,73]} & 56 {\scriptsize[43,68]} & 67 {\scriptsize[49,82]} & 55 {\scriptsize[44,65]} & 59 {\scriptsize[28,86]} & 38 {\scriptsize[0,86]} \\
GPT-5.1 & 59 {\scriptsize[53,66]} & 62 {\scriptsize[55,69]} & 59 {\scriptsize[50,69]} & 53 {\scriptsize[40,66]} & 69 {\scriptsize[52,85]} & 52 {\scriptsize[35,68]} & 64 {\scriptsize[33,89]} & 75 {\scriptsize[54,100]} \\
MiMo v2.5 & 59 {\scriptsize[53,65]} & 61 {\scriptsize[55,68]} & 64 {\scriptsize[55,73]} & 52 {\scriptsize[40,64]} & 67 {\scriptsize[50,81]} & 45 {\scriptsize[30,58]} & 59 {\scriptsize[36,81]} & 50 {\scriptsize[20,100]} \\
GLM-5V Turbo & 59 {\scriptsize[53,65]} & 61 {\scriptsize[55,68]} & 57 {\scriptsize[49,65]} & 53 {\scriptsize[41,65]} & 77 {\scriptsize[62,89]} & 55 {\scriptsize[45,73]} & 68 {\scriptsize[42,90]} & 25 {\scriptsize[0,86]} \\
Claude Opus 4.6 & 58 {\scriptsize[52,65]} & 62 {\scriptsize[55,69]} & 62 {\scriptsize[52,73]} & 61 {\scriptsize[49,71]} & 56 {\scriptsize[39,72]} & 39 {\scriptsize[20,62]} & 59 {\scriptsize[29,84]} & 50 {\scriptsize[20,100]} \\
Qwen 3.6-35B-A3B & 58 {\scriptsize[51,64]} & 59 {\scriptsize[52,66]} & 62 {\scriptsize[54,70]} & 48 {\scriptsize[35,62]} & 62 {\scriptsize[38,83]} & 45 {\scriptsize[32,61]} & 64 {\scriptsize[35,89]} & 62 {\scriptsize[29,100]} \\
Gemma-4-31B & 57 {\scriptsize[51,64]} & 59 {\scriptsize[51,66]} & 64 {\scriptsize[54,72]} & 52 {\scriptsize[40,64]} & 62 {\scriptsize[44,79]} & 42 {\scriptsize[20,63]} & 50 {\scriptsize[13,83]} & 50 {\scriptsize[0,86]} \\
Grok-4.3 & 56 {\scriptsize[49,62]} & 57 {\scriptsize[49,65]} & 62 {\scriptsize[52,72]} & 56 {\scriptsize[45,67]} & 46 {\scriptsize[29,63]} & 42 {\scriptsize[17,59]} & 50 {\scriptsize[15,81]} & 75 {\scriptsize[40,100]} \\
Qwen 3.6-27B & 55 {\scriptsize[49,62]} & 57 {\scriptsize[50,65]} & 61 {\scriptsize[51,70]} & 47 {\scriptsize[36,58]} & 51 {\scriptsize[31,71]} & 48 {\scriptsize[35,62]} & 59 {\scriptsize[25,88]} & 62 {\scriptsize[29,100]} \\
Claude Sonnet 4.6 & 54 {\scriptsize[48,61]} & 56 {\scriptsize[49,63]} & 59 {\scriptsize[49,68]} & 55 {\scriptsize[44,66]} & 54 {\scriptsize[37,69]} & 39 {\scriptsize[19,50]} & 59 {\scriptsize[28,88]} & 25 {\scriptsize[0,86]} \\
Nemotron-Omni & 51 {\scriptsize[45,57]} & 53 {\scriptsize[46,59]} & 56 {\scriptsize[49,63]} & 47 {\scriptsize[32,61]} & 49 {\scriptsize[28,69]} & 39 {\scriptsize[21,49]} & 55 {\scriptsize[26,79]} & 38 {\scriptsize[0,86]} \\
Claude Haiku 4.5 & 49 {\scriptsize[43,55]} & 51 {\scriptsize[44,58]} & 50 {\scriptsize[39,60]} & 44 {\scriptsize[32,56]} & 54 {\scriptsize[37,69]} & 55 {\scriptsize[45,64]} & 50 {\scriptsize[21,79]} & 38 {\scriptsize[0,71]} \\
Mistral Medium 3.5 & 49 {\scriptsize[42,56]} & 52 {\scriptsize[44,59]} & 55 {\scriptsize[44,65]} & 32 {\scriptsize[23,40]} & 64 {\scriptsize[43,82]} & 35 {\scriptsize[19,48]} & 45 {\scriptsize[16,69]} & 75 {\scriptsize[0,100]} \\
Gemma-4-26B-A4B & 49 {\scriptsize[42,55]} & 49 {\scriptsize[42,56]} & 54 {\scriptsize[45,63]} & 39 {\scriptsize[28,52]} & 59 {\scriptsize[43,73]} & 32 {\scriptsize[17,50]} & 45 {\scriptsize[14,76]} & 50 {\scriptsize[20,100]} \\
Qwen 3.5-9B & 47 {\scriptsize[41,53]} & 50 {\scriptsize[43,57]} & 53 {\scriptsize[44,63]} & 36 {\scriptsize[23,48]} & 46 {\scriptsize[25,67]} & 45 {\scriptsize[22,69]} & 50 {\scriptsize[25,71]} & 38 {\scriptsize[20,75]} \\
Gemma-4-E4B (base) & 45 {\scriptsize[39,51]} & 44 {\scriptsize[37,50]} & 48 {\scriptsize[38,57]} & 41 {\scriptsize[29,52]} & 56 {\scriptsize[41,73]} & 32 {\scriptsize[23,48]} & 41 {\scriptsize[15,64]} & 38 {\scriptsize[14,100]} \\
Grok-4.20 & 40 {\scriptsize[35,45]} & 41 {\scriptsize[34,47]} & 41 {\scriptsize[33,48]} & 38 {\scriptsize[28,49]} & 41 {\scriptsize[26,57]} & 52 {\scriptsize[35,65]} & 18 {\scriptsize[0,43]} & 50 {\scriptsize[20,100]} \\
\midrule
Chance & 32.2 & 33.3 & 31.8 & 32.8 & 29.7 & 34.9 & 30.7 & 40.6 \\
\bottomrule
\end{tabularx}
\end{table*}

\paragraph{Controls on the audited subset.}
\Cref{tab:controls-audited} repeats Tab.~3 on the 254 audited questions; the conclusions of \main{5} are unchanged.

\begin{table*}[t]
\centering
\caption{Input controls on VisionQ-Bench (audited subset, 254 questions). $\Delta$: paired change from the standard run (\colorbox{up!35}{green} up, \colorbox{down!35}{red} down; * 95\% paper-bootstrap interval excludes zero; --: n/a). Bold: best in the row. \emph{Same crop} at random: 13.8\%. $^\dagger$On the 175 questions with a reference image.}
\label{tab:controls-audited}
\footnotesize
\setlength{\tabcolsep}{1pt}
\begin{tabularx}{\textwidth}{@{}l>{\raggedleft\arraybackslash}X>{\centering\arraybackslash}X>{\raggedleft\arraybackslash}X>{\centering\arraybackslash}X>{\raggedleft\arraybackslash}X>{\centering\arraybackslash}X>{\raggedleft\arraybackslash}X>{\centering\arraybackslash}X>{\raggedleft\arraybackslash}X>{\centering\arraybackslash}X>{\raggedleft\arraybackslash}X>{\centering\arraybackslash}X@{}}
\toprule
 & \multicolumn{2}{c}{\scriptsize\textbf{Gemma-4-E4B}} & \multicolumn{2}{c}{\scriptsize\textbf{Gemma-4-26B-A4B}} & \multicolumn{2}{c}{\scriptsize\textbf{Gemma-4-31B}} & \multicolumn{2}{c}{\scriptsize\textbf{MiMo v2.5}} & \multicolumn{2}{c}{\scriptsize\textbf{GPT-5.3-codex}} & \multicolumn{2}{c}{\scriptsize\textbf{Gemini 3.1 Pro}} \\
\cmidrule(lr){2-3}\cmidrule(lr){4-5}\cmidrule(lr){6-7}\cmidrule(lr){8-9}\cmidrule(lr){10-11}\cmidrule(lr){12-13}
Condition & Acc. & $\Delta$ & Acc. & $\Delta$ & Acc. & $\Delta$ & Acc. & $\Delta$ & Acc. & $\Delta$ & Acc. & $\Delta$ \\
\midrule
Standard & 46.1 & -- & 55.5 & -- & 59.4 & -- & 60.6 & -- & 65.4 & -- & \textbf{74.4} & -- \\
\midrule
\multicolumn{13}{@{}l}{\textbf{E1: option order}} \\
Mean over orders & 44.3 & \cellcolor{down!10}$-$1.8 & 55.5 & \cellcolor{up!6}+0.0 & 61.5 & \cellcolor{up!10}+2.0 & 62.2 & \cellcolor{up!9}+1.6 & 63.6 & \cellcolor{down!9}$-$1.7 & \textbf{72.3} & \cellcolor{down!10}$-$2.1 \\
Plurality vote & 35.0 & \cellcolor{down!28}$-$11.0* & 50.4 & \cellcolor{down!16}$-$5.1 & 62.2 & \cellcolor{up!12}+2.8 & 59.1 & \cellcolor{down!9}$-$1.6 & 61.4 & \cellcolor{down!14}$-$3.9 & \textbf{71.3} & \cellcolor{down!12}$-$3.1 \\
Same crop in all orders & 19.3 & -- & 42.9 & -- & 52.8 & -- & 46.1 & -- & 52.8 & -- & \textbf{66.1} & -- \\
\midrule
\multicolumn{13}{@{}l}{\textbf{E2: criterion}} \\
Criterion removed & 41.7 & \cellcolor{down!15}$-$4.3 & 56.3 & \cellcolor{up!8}+0.8 & 57.5 & \cellcolor{down!10}$-$2.0 & 58.7 & \cellcolor{down!10}$-$2.0 & 57.9 & \cellcolor{down!21}$-$7.5* & \textbf{66.9} & \cellcolor{down!21}$-$7.5* \\
\midrule
\multicolumn{13}{@{}l}{\textbf{E3: images}} \\
Blank tiles & 32.3 & \cellcolor{down!34}$-$13.8* & 31.1 & \cellcolor{down!55}$-$24.4* & 29.1 & \cellcolor{down!60}$-$30.3* & 28.0 & \cellcolor{down!60}$-$32.7* & 33.1 & \cellcolor{down!60}$-$32.3* & 31.9 & \cellcolor{down!60}$-$42.5* \\
Mismatched composite & 35.4 & \cellcolor{down!27}$-$10.6* & 33.1 & \cellcolor{down!51}$-$22.4* & 33.1 & \cellcolor{down!59}$-$26.4* & 30.3 & \cellcolor{down!60}$-$30.3* & 28.7 & \cellcolor{down!60}$-$36.6* & 33.1 & \cellcolor{down!60}$-$41.3* \\
\midrule
\multicolumn{13}{@{}l}{\textbf{E4: restored inputs}} \\
Full question & 44.9 & \cellcolor{down!8}$-$1.2 & 54.7 & \cellcolor{down!8}$-$0.8 & 60.6 & \cellcolor{up!8}+1.2 & 62.2 & \cellcolor{up!9}+1.6 & 65.0 & \cellcolor{down!7}$-$0.4 & \textbf{72.0} & \cellcolor{down!11}$-$2.4 \\
Full question + ref.$^\dagger$ & 39.4 & \cellcolor{down!20}$-$6.9 & 58.9 & \cellcolor{up!22}+8.0 & 67.4 & \cellcolor{up!28}+10.9* & 65.1 & \cellcolor{up!23}+8.6 & 70.3 & \cellcolor{up!24}+9.1* & \textbf{77.7} & \cellcolor{up!16}+5.1 \\
\bottomrule
\end{tabularx}
\end{table*}

\paragraph{Image-quality baselines.}
Each baseline scores every candidate crop on its own and picks the crop with the highest predicted quality.
Sharpness is the variance of a $3{\times}3$ Laplacian on the grey image; MUSIQ~\cite{musiq}, TOPIQ-NR~\cite{topiq} and CLIP-IQA+~\cite{clipiqa} use their KonIQ-trained weights from the IQA-PyTorch toolbox.
\Cref{tab:s-iqa} gives the results.

\begin{table}[t]
\centering
\caption{No-reference image-quality baselines on the 309 scored questions (chance 32.2\%; strongest judge 68.6\%).}
\label{tab:s-iqa}
\footnotesize
\begin{tabular}{@{}lc@{}}
\toprule
\textbf{Pick the crop with the highest} & \textbf{Accuracy (\%)} \\
\midrule
Laplacian sharpness & 36.2 {\scriptsize[30,43]} \\
MUSIQ & 40.1 {\scriptsize[34,46]} \\
TOPIQ-NR & 40.8 {\scriptsize[34,47]} \\
CLIP-IQA+ & 41.7 {\scriptsize[35,49]} \\
\bottomrule
\end{tabular}
\end{table}

\paragraph{Recognition of published figures.}
Each scored question was asked twice in one run with the benchmark's renderer, prompt and parser: with the original crops and with every crop mirrored left to right.
Mirroring keeps the relative quality of the outputs but breaks an exact match with the published figure (\cref{tab:s-recog}).

\begin{table}[t]
\centering
\caption{Accuracy (\%) with original and mirrored crops on the 309 scored questions. \emph{Change}: mirrored minus original, paired by question, with its 95\% paper-bootstrap interval. \emph{Same}: share of questions answered with the same crop in both runs.}
\label{tab:s-recog}
\footnotesize
\setlength{\tabcolsep}{3pt}
\begin{tabular}{@{}lcccc@{}}
\toprule
\textbf{Judge} & \textbf{Original} & \textbf{Mirrored} & \textbf{Change} & \textbf{Same} \\
\midrule
Gemini 3.1 Pro & 69.3 & 68.9 & $-$0.3 {\scriptsize[$-$3.5,+2.8]} & 89.0 \\
GPT-5.3-codex & 60.8 & 61.8 & +1.0 {\scriptsize[$-$3.8,+5.6]} & 79.3 \\
Gemma-4-31B & 57.0 & 58.9 & +1.9 {\scriptsize[$-$3.4,+7.1]} & 71.5 \\
\bottomrule
\end{tabular}
\end{table}

\section{Human study}
\label{sec:s-human}

\paragraph{Sample.}
The 150 questions include every Relation (31), Prompt Match (22) and Scene Layout (8) question, so that the small axes are covered, and 89 questions from Object Form, Reference Fidelity and Image Appearance chosen to spread over as many papers as possible.
Of these, 126 pass the option-label audit (\cref{sec:s-items}) and are analysed in \main{5}.
On the other 24, the raters agree with the key on 29\% and 4\%.

\paragraph{Interface.}
Raters worked in a web interface with accounts issued for the study.
Each question shows the candidate crops with letters and the full sanitised question; each rater sees the options in a private order, so letters differ between raters.
Raters choose one output or ``cannot tell / tie'' and rate their confidence from 1 (low) to 3 (high).
When the source figure has a reference image, a second screen adds it above the candidates and asks again; the first answer is locked before the second screen appears.
Generation prompts are not available in our data, so Prompt Match questions are judged without them.

\paragraph{Instructions.}
Raters received the following instructions:
\begin{quote}\footnotesize
For each question, choose the output that best satisfies the stated visual criterion, using only what is visible in the displayed crops. Judge the criterion in the question, not which image looks most attractive overall. Choose \emph{Cannot tell / tie} when the visible evidence does not let you make a reliable distinction, or when candidates appear equally good for the stated criterion. [\ldots] Some items have a second screen that adds reference or context tiles above the candidate images. Make a fresh judgment using all visible material; your first answer is locked. [\ldots] Do not search for or read the source papers, and do not discuss answers with other annotators before submitting.
\end{quote}

\paragraph{Participants and consent.}
The two raters are CV-literate, are not the annotators who linked rows to claims, did not build the answer key, participated voluntarily and gave consent.
The study recorded only their answers and confidence ratings.
No ethics review was sought, because participants judged only published figure crops and no personal or sensitive data were collected.

\paragraph{Further results.}
The sample holds every Relation, Prompt Match and Scene Layout item and 89 items from the three larger axes; generation prompts are not available, so Prompt Match items are judged without them.
The second rater agrees with the key on 68.2\% of the questions it decides.
Both raters decide on 88 questions. The main text compares them on this set. There, Rater~1 chooses the key on 55 (62.5\%) and Rater~2 on 60 (68.2\%), and the two choose the same crop on 62 (70.5\%). Chance on these 88 questions, the mean of one over the number of options, is 32.7\%.
The reference image raises agreement for both raters; these second judgments followed the first on the same items, so the change may also include repeat exposure.
On the same 126 questions, Gemini 3.1 Pro reaches 71.4\% and GPT-5.3-codex 61.9\% in the controls' standard runs; the comparison is descriptive, since the raters read the full question in a private option order and could abstain.

\paragraph{Licensing and ethics.}
The release is hosted at \url{https://huggingface.co/datasets/visionq-review/VisionQ} (the annotated pool of 3{,}911 comparisons from 810 papers with all 22{,}148 crops, VisionQ-Bench and VisionQ-MCQ), \url{https://huggingface.co/visionq-review/VisionQ-Judge} (VisionQ-Judge adapters for the three seeds) and \url{https://github.com/ReML-AI/visionq} (code and the predictions behind every table).
Annotations are released under CC BY 4.0, while figure crops remain the property of their authors and publishers and are distributed for non-commercial research with crop coordinates, source references and a removal procedure for authors who opt out.
No ethics review was sought; we considered the task minimal risk because participants judged only published figure crops using study accounts that we issued, and the study recorded only their answers and confidence ratings, with no personal or sensitive data.
\section{VisionQ-Judge}

\label{sec:s-judge}

\paragraph{VisionQ-MCQ.}
VisionQ-MCQ is built from the annotated pool without language-model calls.
A record is kept when its claim names a winner, its comparison mode places competing methods side by side, and the winner's crop can be identified: the winner name from the claim is matched against the crop labels, with a fall-back to the crop labelled as the authors' own method (``ours'').
Up to three distractors are chosen from the other methods of the same row; crops labelled as ground truth, input or reference are excluded, and each method contributes one crop.
This gives 4{,}524 questions from 513 papers.
VisionQ-Judge is trained and evaluated on VisionQ-MCQ only. VisionQ-Bench is kept for judges used as released, so that every judge on the leaderboard is scored without task-specific training, and its 309 questions are too few for a paper-disjoint training split.

\paragraph{Splits.}
Each leaf is capped at 300 questions (4{,}524 to 3{,}584).
One split was fixed before any training run: every paper lies on exactly one side, giving 2{,}687 training, 359 validation and 538 test questions from 239, 79 and 168 papers.
The split is stratified by leaf: all 46 leaves in the pool have a training paper, every leaf with at least two papers has a test paper (38 leaves) and every leaf with at least three has a validation paper (31), and each side's axis mix is within 1.5 points of the pool's.

\paragraph{Prompt.}
The crops are tiled into one image with letter badges (one row for two options, two rows otherwise; 512\,px per tile).
The text, used identically in training and evaluation and for every judge compared on VisionQ-MCQ, is:
\begin{quote}\footnotesize\ttfamily\raggedright
Context:\\
Row context: \{row name\}. Source paper caption: \{caption, cut at 400 characters\}\\[3pt]
\{question\}\\[3pt]
Select from the following choices:\\
(A) panel A of the image\\
(B) panel B of the image \ldots\\[3pt]
Look carefully at every option before deciding; do not default to any particular position. Use the visual evidence.
\end{quote}

\paragraph{Symmetric evidence pairs.}
Pairs are produced deterministically from the stored claim, criterion, winner and crops, without calls to a language model.
The evidence $e$ is the author claim with the winner's name replaced by its letter (\eg ``image C method produces more accurate detailed geometry''), or the claimed visual attribute when the claim cannot be normalised.
For a question with winner $w$, each wrong option $r$ gives one pair with chosen response ``($w$)'' followed by $e$ and rejected response ``($r$)'' followed by $e$ with every mention of $w$'s letter replaced by $r$'s.
Each training question is rendered in every cyclic order of its options, and both the answer letter and the letters inside $e$ are remapped so that they follow the winner to its new position.
The 2{,}687 training questions become 7{,}083 rendered questions and 13{,}394 pairs.
The \emph{one reordered copy} variant renders each question in its original and one other order (8{,}786 pairs).

\paragraph{Training.}
We fine-tune \texttt{google/gemma-4-E4B-it} with DPO~\cite{dpo} (TRL 1.7.0, Transformers 5.12.1, PyTorch 2.8.0) and LoRA~\cite{lora} adapters (rank 16, $\alpha=32$, dropout 0.05) on the attention and MLP projections of the language decoder and the attention projections of the vision encoder.
Settings: $\beta=0.3$, learning rate $2{\times}10^{-5}$ with 5\% warm-up, one epoch, batch size 1 with 8 accumulation steps, half precision, at most 1{,}024 tokens and 512\,px per image.
The three seeds (42, 43, 44) share the data and split.

\paragraph{Per-seed results.}
\Cref{tab:s-seeds} gives every run on the test and validation splits, evaluated on one GPU type with greedy decoding.

\begin{table}[t]
\centering
\caption{Accuracy and share of answers on the last option (\%) per run on VisionQ-MCQ. In the key, the answer is last on 38.8\% of test questions.}
\label{tab:s-seeds}
\footnotesize
\setlength{\tabcolsep}{3pt}
\begin{tabular}{@{}lcccc@{}}
\toprule
& \multicolumn{2}{c}{\textbf{Test (538)}} & \multicolumn{2}{c}{\textbf{Validation (359)}} \\
\cmidrule(lr){2-3}\cmidrule(lr){4-5}
\textbf{Run} & Acc. & Last & Acc. & Last \\
\midrule
Base model & 50.4 & 50.7 & 44.0 & 44.8 \\
\midrule
VisionQ-Judge, seed 42 & 61.3 & 42.9 & 54.0 & 38.4 \\
VisionQ-Judge, seed 43 & 60.2 & 45.5 & 54.0 & 40.7 \\
VisionQ-Judge, seed 44 & 61.3 & 43.1 & 54.6 & 40.1 \\
\midrule
One reordered copy, seed 42 & 58.0 & 46.5 & 51.0 & 39.3 \\
One reordered copy, seed 43 & 58.2 & 49.6 & 52.6 & 46.8 \\
One reordered copy, seed 44 & 56.9 & 46.3 & 54.0 & 38.4 \\
\bottomrule
\end{tabular}
\end{table}

\paragraph{Results in detail.}
Test accuracy over the three seeds is 61.3, 60.2 and 61.3\%. On the 329 test questions whose answer is not last, accuracy rises from 40.7\% to 54.9\% (+14.2, CI +5.5 to +22.0), and on the 209 whose answer is last from 65.6\% to 70.5\% (+4.9, CI $-$1.2 to +11.7) (Fig.~5).
Counting answers over orders does not help the base model: it picks the same crop in every cyclic order on only 26.8\% of questions, and a plurality vote falls to 31.4\%.
Averaging the letter log-probabilities differs from the one-call VisionQ-Judge by +0.7 points (CI $-$5.3 to +6.7) and adds 3.8 points to VisionQ-Judge (CI $-$0.1 to +7.9).
One reordered copy of each question uses 8{,}786 pairs; rendering every cyclic order (2{,}687 questions become 7{,}083, giving 13{,}394 pairs) adds +1.7 points on validation over one copy (CI $-$0.6 to +3.9) and raises the share of questions with the same crop in every order from 46.2\% to 50.2\%.
\paragraph{Inference-time alternatives.}
For \main{6}, we read the next-token distribution after the answer prefix ``('' and keep the log-probabilities of the option letters, for the question in every cyclic order; on one GPU, this readout matches greedy generation on all 30 questions we checked.
\emph{Log-probability averaging} maps each order's letters back to the original options, averages the log-probabilities per option over all cyclic orders and picks the highest; this costs one call per order (2.8 on average).
\emph{PriDe}~\cite{wang2023large} estimates a prior over letter positions without labels on the validation split, separately for two, three and four options, from each question's log-probabilities averaged over its cyclic orders; at test time it subtracts the log-prior from a single call's log-probabilities.
The estimated prior is close to uniform (four options: 0.29, 0.20, 0.24, 0.27).
Tab.~5c gives the results.

\paragraph{Sensitivity to the GPU type.}
Gemma-4-E4B runs in half precision, and its greedy answers depend on the GPU type.
Between our two GPU types, the base model gives the same answer on 421 of the 538 test questions and VisionQ-Judge (seed 42) on 436.
Absolute accuracy shifts (base model 50.4\% against 47.0\%, VisionQ-Judge 61.0\% against 57.4\%), but the gain from training is the same (+10.6 against +10.3 points).
We therefore run both conditions of every paired comparison on one GPU type: the main results on one type, and the probability readout (Tab.~5c) and the E2 comparison below on the other.

\paragraph{Criterion removed (E2) on VisionQ-MCQ.}
With the question replaced by ``Which output is better overall?'' and the caption kept, accuracy changes by +3.7 points for the base model (95\% interval $-$0.8 to +8.3) and by +0.4 for VisionQ-Judge ($-$2.4 to +3.1).

\FloatBarrier
\renewcommand{\topfraction}{0.9}
\renewcommand{\bottomfraction}{0.8}
\renewcommand{\dbltopfraction}{0.97}
\renewcommand{\textfraction}{0.1}
\renewcommand{\floatpagefraction}{0.8}
\renewcommand{\dblfloatpagefraction}{0.75}
\widowpenalty=10000
\clubpenalty=10000
\makeatletter
\setlength{\@dblfptop}{0pt}
\setlength{\@dblfpsep}{10pt plus 1fil}
\setlength{\@dblfpbot}{0pt plus 1fil}
\makeatother
\section{Criterion Sensitivity, Position Bias and Visual Grounding of the Judges}
\label{sec:s-attention}

\begin{table*}[t]
\centering
\caption{Criterion sensitivity. To test whether a judge uses the named criterion, we change only the question and measure how far its attention and its answers move from those under the named criterion (538 VisionQ-MCQ test questions). A rewording gives the level expected when the criterion is unchanged, and a different image the level expected when the whole input changes; a criterion-sensitive judge would place the different-criterion rows near the latter. The bottom block applies the same test to visual question answering (GQA~\cite{hudson2019gqa}, 200 images), where the question determines where a model looks; there, the reworded question is a paraphrase written for this study. \emph{Overlap}: intersection over union of the most-attended quarters of the two attention maps~\cite{vea} (\%). \emph{Changed}: answers that differ (\%). \emph{Acc.}: accuracy (\%) from one forward pass per question (\cref{tab:s-seeds} uses greedy decoding). $\Delta$: paired change in accuracy from the named criterion (\colorbox{up!35}{green} up, \colorbox{down!35}{red} down; * 95\% question-bootstrap interval excludes zero; --: n/a). Bold: best in the row. \emph{No sinks}: overlap without sink tokens~\cite{kang2025visualsink}. \emph{No criterion}: ``Which output is better overall?''. \emph{Different criterion}: drawn at random from another axis of the vocabulary (\cref{sec:s-vocab}), or taken from a test question of a different paper.}
\label{tab:s-stability}
\footnotesize
\setlength{\tabcolsep}{1.5pt}
\begin{tabular}{@{}lccccc@{\hspace{5pt}}ccccc@{\hspace{5pt}}ccccc@{}}
\toprule
& \multicolumn{5}{c}{\scriptsize\textbf{Gemma-4-E2B}} & \multicolumn{5}{c}{\scriptsize\textbf{Gemma-4-E4B}} & \multicolumn{5}{c}{\scriptsize\textbf{VisionQ-Judge}} \\
\cmidrule(lr){2-6}\cmidrule(lr){7-11}\cmidrule(lr){12-16}
Question shown & Overlap & No sinks & Changed & Acc. & $\Delta$ & Overlap & No sinks & Changed & Acc. & $\Delta$ & Overlap & No sinks & Changed & Acc. & $\Delta$ \\
\midrule
\multicolumn{16}{@{}l}{\textbf{VisionQ-MCQ test set (538 questions)}} \\
Named criterion & 100.0 & 100.0 & 0.0 & 43.3 & -- & 100.0 & 100.0 & 0.0 & 49.1 & -- & 100.0 & 100.0 & 0.0 & \textbf{58.9} & -- \\
Reworded & 88.8 & 77.4 & 14.5 & 43.5 & \cellcolor{up!6}+0.2 & 80.5 & 67.2 & 15.4 & 52.4 & \cellcolor{up!13}+3.3* & 78.1 & 65.9 & 9.7 & \textbf{62.5} & \cellcolor{up!13}+3.5* \\
No criterion & 82.3 & 65.9 & 21.7 & 44.6 & \cellcolor{up!9}+1.3 & 75.8 & 57.9 & 21.6 & 53.2 & \cellcolor{up!15}+4.1* & 72.7 & 57.1 & 14.7 & \textbf{60.2} & \cellcolor{up!9}+1.3 \\
Different criterion (vocabulary) & 79.6 & 62.3 & 29.6 & 45.9 & \cellcolor{up!11}+2.6 & 71.1 & 51.0 & 27.1 & 47.4 & \cellcolor{down!10}$-$1.7 & 66.6 & 50.0 & 20.6 & \textbf{55.6} & \cellcolor{down!13}$-$3.3 \\
Different criterion (question) & 80.8 & 64.8 & 28.1 & 41.3 & \cellcolor{down!10}$-$2.0 & 74.8 & 56.9 & 26.6 & 49.4 & \cellcolor{up!7}+0.4 & 70.5 & 56.3 & 16.0 & \textbf{57.8} & \cellcolor{down!8}$-$1.1 \\
Different image & 61.3 & 30.3 & -- & -- & -- & 42.8 & 24.2 & -- & -- & -- & 30.5 & 19.6 & -- & -- & -- \\
\midrule
\multicolumn{16}{@{}l}{\textbf{GQA (200 images)}} \\
Reworded question & \multicolumn{5}{c}{76.1} & \multicolumn{5}{c}{78.3} & \multicolumn{5}{c}{76.9} \\
Different question & \multicolumn{5}{c}{28.9} & \multicolumn{5}{c}{28.7} & \multicolumn{5}{c}{26.8} \\
Different image & \multicolumn{5}{c}{22.2} & \multicolumn{5}{c}{16.9} & \multicolumn{5}{c}{16.8} \\
\bottomrule
\end{tabular}
\end{table*}

In the main text, we evaluate judges by their accuracy.
We find that they keep most of their accuracy when the criterion is removed and change their answers when the options are reordered (Sec.~5).
Training on every option order raises accuracy (Sec.~6).
In this section, we investigate why.
These observations correspond to three weaknesses that prior work reports for language-model judges and vision-language models.
Text-only judges confuse the criteria that they are asked to apply~\cite{hu2024criteria}.
Judges and multiple-choice models prefer an option for its position or its label rather than for its content~\cite{mtbench,wang2023large}.
Vision-language models can answer correctly without relying on the relevant image content~\cite{wrongReasons,fpvg}.
We therefore study the three corresponding properties of a judge.
\emph{Criterion sensitivity} is how much its attention and answers change when we change the criterion.
\emph{Position bias} is how often its answer stays at an option position when we move the outputs.
\emph{Visual grounding} is how much its answer depends on the output images.
A judge that compares outputs on the named criterion should be sensitive to the criterion, free of position bias and grounded in the images.
We measure the three properties on Gemma-4-E2B, Gemma-4-E4B and VisionQ-Judge, and then compare VisionQ-Judge with its base model to see which properties training changes.
Our results point to an explanation for each of these observations.
These judges keep most of their accuracy without the criterion because they are only weakly sensitive to it.
The base models change their answers when the options are reordered because they have a strong position bias.
Training reduces this bias and strengthens visual grounding, but it does not make the judge more sensitive to the criterion.

\paragraph{Setup.}
We use the 538 VisionQ-MCQ test questions and the prompt of \cref{sec:s-judge}, with 280 image tokens per image.
We use two methods.
The first is the attention map: the attention from the answer position to the image tokens, averaged over heads and over the visual-grounding layers selected as in VEA~\cite{vea}.
To compare two attention maps, we take the most-attended quarter of each and compute their intersection over union, which we call the \emph{overlap}.
The second is attention knockout~\cite{geva2023dissecting,crossmodalflow}.
Following Geva et al.~\cite{geva2023dissecting} and Zhang et al.~\cite{crossmodalflow}, we block one part of the prompt from attending to another in a window of nine consecutive layers.
We then measure the relative drop in the probability of the answer.
In the attention maps, about one tenth of the image tokens receive most of the attention under every prompt (10 to 13\% of the tokens receive 86 to 91\% of the attention).
Kang et al.~\cite{kang2025visualsink} report a related phenomenon, visual attention sinks.
We call these tokens sink tokens, define them by attention alone, and report every overlap with and without them.
In this setup, accuracy comes from one forward pass per question.
This accuracy differs from the main results: 49.1 and 58.9\% for the base model and VisionQ-Judge, against 50.4 and 61.0\% in \main{6}.
A change of GPU type alone shifts accuracy by up to 3.6 points (\cref{sec:s-judge}).
The change without the criterion also differs slightly from the E2 run in \cref{sec:s-judge} (+4.1 and +1.3 points here, against +3.7 and +0.4).
Neither run shows a loss without the criterion.
We therefore compare conditions only within this setup, and every change we report is paired.

\begin{table}[t]
\centering
\caption{Position bias and visual grounding (538 VisionQ-MCQ test questions; E2B knockout: 100). To test whether an answer is tied to a position, we move the outputs by one position with the letters fixed (\emph{Shifted}) and record whether the answer stays at its position or follows the output it chose before; with three or four options it can also move elsewhere. To test whether an answer depends on the images, we replace every output by a grey tile (\emph{Greyed tiles}: paired change in accuracy; \colorbox{down!35}{red} down; * 95\% question-bootstrap interval excludes zero; chance is 39\%) and block attention paths (\emph{Knockout}: attention knockout~\cite{geva2023dissecting,crossmodalflow}; ``image to X'' is the largest relative drop in the probability of the answer when X is blocked from attending to the image).}
\label{tab:s-use}
\footnotesize
\setlength{\tabcolsep}{3pt}
\begin{tabular}{@{}lccc@{}}
\toprule
& \textbf{E2B} & \textbf{E4B} & \textbf{VisionQ-Judge} \\
\midrule
\multicolumn{4}{@{}l}{\textbf{Position bias}} \\
Shifted: stays at the position (\%) & 73.8 & 53.5 & 27.5 \\
Shifted: follows the output (\%) & 16.5 & 36.8 & 58.2 \\
Shifted, no letters: stays (\%) & 78.4 & 53.5 & 31.2 \\
\midrule
\multicolumn{4}{@{}l}{\textbf{Visual grounding}} \\
Greyed tiles (points) & \cellcolor{down!21}$-$7.1* & \cellcolor{down!29}$-$11.2* & \cellcolor{down!48}$-$19.9* \\
Knockout, image to option list (\%) & 25.7 & 21.5 & 12.9 \\
Knockout, image to answer (\%) & 4.1 & 1.8 & 15.6 \\
Knockout, image to criterion (\%) & 3.7 & 0.9 & 1.6 \\
\bottomrule
\end{tabular}
\end{table}

\paragraph{Validation on visual question answering.}
Before we study judging, we apply the two methods to visual question answering on the same three models.
This verifies that the methods reproduce prior findings, and it shows how attention behaves when the text does determine where a model must look.
We make three observations.
First, attention localises the evidence.
We use Visual-CoT~\cite{shao2024visualcot}, which annotates the image region that contains the evidence.
We select the visual-grounding layers on 100 of its questions as in VEA~\cite{vea} and test on 800 other questions.
The attention maps of all three models separate the evidence region from the rest of the image (AUROC 0.84, against 0.5 by chance), in line with prior work~\cite{vea,whereToLook}.
Second, attention follows the question.
On 200 GQA images~\cite{hudson2019gqa}, a paraphrase of the question, with the same meaning in different words, largely preserves the attention map (overlap 76 to 78\%).
A different question about the same image moves it almost as much as a different image does (27 to 29\% against 17 to 22\%, \cref{tab:s-stability}, bottom, and \cref{fig:s-swap}, right).
Third, the question reads the image.
Blocking the question from attending to the image lowers the probability of the answer by 25 to 35\%, compared with 3 to 6\% for random text tokens.
This drop reproduces the flow from the image to the question reported by Zhang et al.~\cite{crossmodalflow}.

These observations hold equally for VisionQ-Judge and its base model, so training for judging does not change this behaviour.
They also give the reference against which we compare judging below.

\begin{figure*}[t]
\centering
\includegraphics[width=\textwidth]{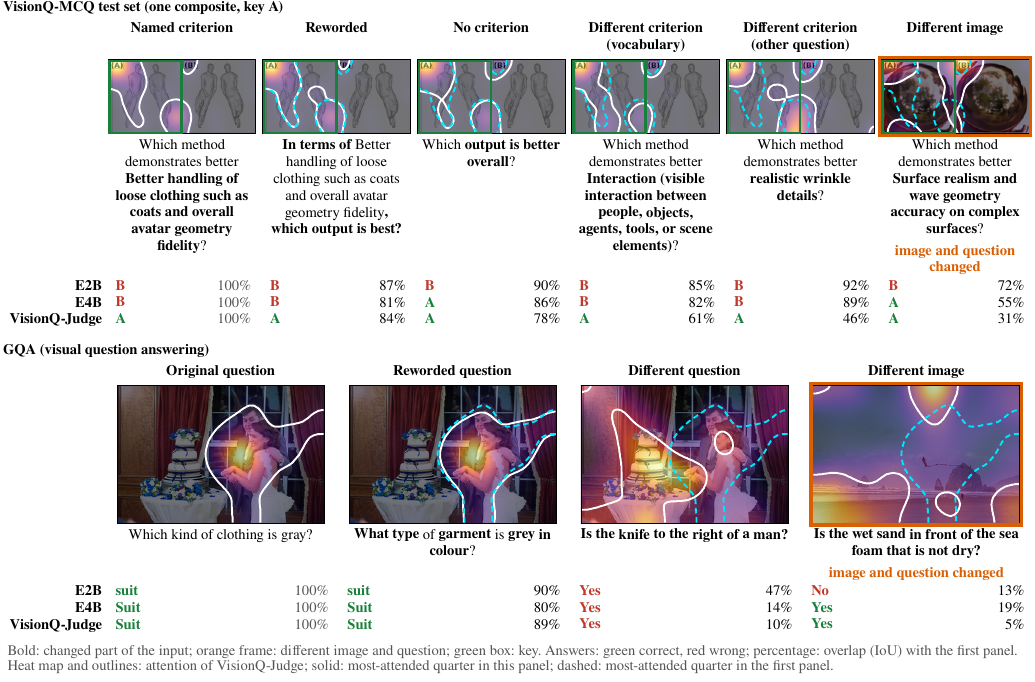}
\caption{The criterion-sensitivity test on one example. We change only the question shown to the judge (changed part in bold) and compare the attention of VisionQ-Judge (heat map) with its attention under the original question. Top: on a VisionQ-MCQ composite, the most-attended region (solid outline) stays close to that under the named criterion (dashed outline) when the criterion is reworded or removed (overlap 84 and 78\%). The overlap falls when the criterion is replaced (61 and 46\%) and is lowest when the image is replaced (31\%). Bottom: on a GQA image, the region is preserved under a paraphrase (89\%) but moves when a different question is asked (10\%). The numbers below each panel give the answer of each judge and the overlap of its two regions. Both examples are the items at the median overlap over judges (rank 269 of 538 and 101 of 200); the averages are in \cref{tab:s-stability}.}
\label{fig:s-conditions}
\end{figure*}

\begin{figure*}[t]
\centering
\includegraphics[width=\textwidth]{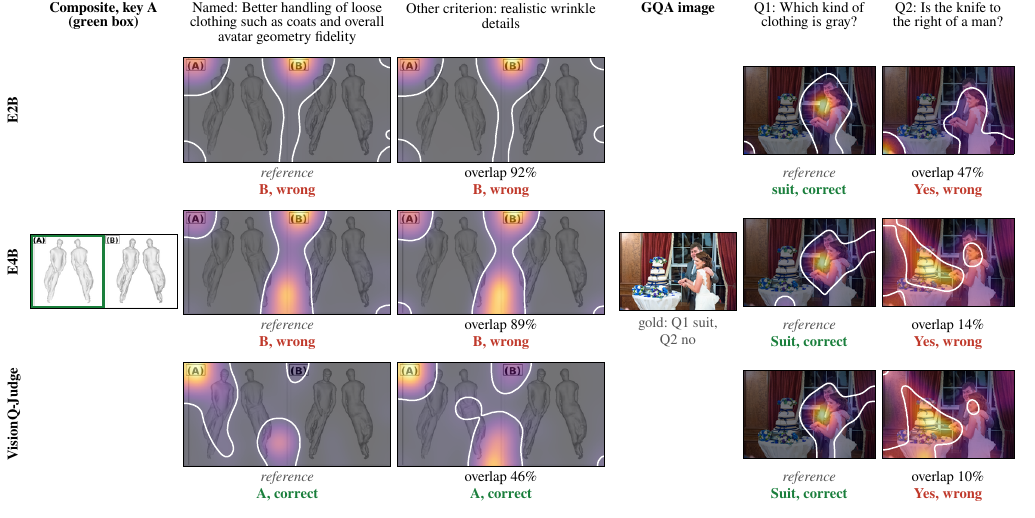}
\caption{Attention maps of the three judges for the examples of \cref{fig:s-conditions}. Left: on the composite, the most-attended region (white outline) under a different criterion overlaps with the region under the named criterion by 92, 89 and 46\% for Gemma-4-E2B, Gemma-4-E4B and VisionQ-Judge. Right: on the GQA image, a different question moves the region for all three judges (overlap 47, 14 and 10\%). The averages over all items are in \cref{tab:s-stability}.}
\label{fig:s-swap}
\end{figure*}

\paragraph{Criterion sensitivity.}
In the main text, judges keep most of their accuracy when the criterion is removed, which suggests that they make little use of it.
Prior work on visual question answering reports the opposite for a question.
The attention of a model moves to the region that the question asks about~\cite{vea,whereToLook}, and the question tokens collect the image information before the answer is produced~\cite{crossmodalflow}.
Both findings hold for our three models, as shown above.
If a judge used the criterion in the same way, a different criterion would move its attention and change its answers, and the correct criterion would improve its accuracy.
We test this prediction by keeping the composite and replacing the named criterion with a different one.
We use two kinds of different criterion.
The first is drawn at random from another axis of our vocabulary (\cref{sec:s-vocab}) and given by name and definition.
The second is the criterion of a test question from a different paper.
We compare the result with two references: a rewording of the original question, which should change little, and a different image, which should change a lot (\cref{tab:s-stability}).
\Cref{fig:s-conditions} shows the input and output of each condition on one example.

For VisionQ-Judge, the attention maps under a different criterion overlap by 67 to 71\% with the map under the named criterion.
This overlap is 8 to 12 points below a rewording (78\%) and far above a different image (31\%).
The other two judges show the same ordering, with and without sink tokens, and \cref{fig:s-swap} shows the attention maps for one example.
The attention therefore changes little across criteria.
This small change contrasts with visual question answering, where a different question lowers the overlap to 27 to 29\%.
We suggest that this difference comes from what a criterion is.
A criterion describes a property of the whole output, not a place in the image, so a judge has to look at every option whichever criterion is named.

The answers change more than the attention.
A different criterion changes 16 to 21\% of the answers of VisionQ-Judge, about twice as many as a rewording.
However, naming the correct criterion does not improve accuracy.
Accuracy with the named criterion is within 3.3 points of that with a different criterion, and never above that with no criterion.
No judge reaches its highest accuracy with the named criterion.
A rewording even raises accuracy by 3.3 and 3.5 points for Gemma-4-E4B and VisionQ-Judge, and both intervals exclude zero.
Attention knockout gives the same picture.
Blocking the criterion sentence from attending to the image lowers the probability of the answer by at most 3.7\% (\cref{tab:s-use}).
This drop is no larger than for random text tokens, and far below the 25 to 35\% for a question in visual question answering.
We conclude that these three judges are only weakly sensitive to the criterion.
This weak sensitivity explains why they keep most of their accuracy without it.

\begin{figure*}[t]
\centering
\includegraphics[width=\textwidth]{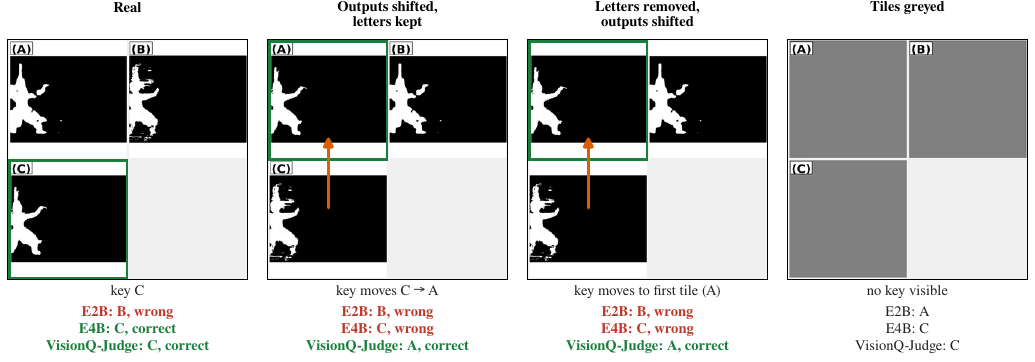}
\caption{Position bias and visual grounding on one test question. We show the composite as given to the judges, with the outputs shifted by one position, with the letters also removed, and with every output replaced by a grey tile. The green box marks the key and the arrow shows where it moves. When the outputs are shifted, the two base models keep their answer at the same position, whereas VisionQ-Judge follows the key; with grey tiles no key remains. The example is illustrative; the rates over all questions are in \cref{tab:s-use}.}
\label{fig:s-position}
\end{figure*}

\paragraph{Position bias.}
In the main text, judges change their answers when the options are reordered.
Language models are known to prefer certain options in multiple-choice questions, and Zheng et al.~\cite{wang2023large} trace this preference to the option letters.
In our setting, the preference could therefore come from the letter of an option or from the position of its output in the image.
We separate these two sources.
We first shift the outputs by one position and keep the letters fixed (\cref{tab:s-use,fig:s-position}).
An answer can then stay at the same position or follow the output that the judge chose before.
Gemma-4-E2B and Gemma-4-E4B mostly stay at the same position (74 and 54\% of answers), while VisionQ-Judge mostly follows the output (58\%).
We then remove the letters from the image and name the options by position.
The result is nearly the same (78, 54 and 31\% of answers stay).
The two base models therefore have a strong bias towards the position of an option, and this bias does not depend on the letters.
This bias explains why their answers change when the options are reordered.

\paragraph{Visual grounding.}
In the main text, judges fall to near chance when the images are blanked, so they need the images.
This result does not show how the images are used.
In visual question answering, Zhang et al.~\cite{crossmodalflow} show that the image information first flows to the question tokens and then to the answer position.
In judging, the image could reach the answer through the criterion sentence, through the option list, or directly.
We first repeat the blank-image control on this test set by replacing every output with a grey tile, an occlusion test~\cite{zeiler2014visualizing}.
Accuracy drops by 7 to 20 points, and all three judges fall to chance (39\%, \cref{tab:s-use}).
We then use attention knockout to block each route in turn.
Blocking the option list from attending to the image lowers the probability of the answer by 13 to 26\%, whereas blocking the criterion sentence lowers it by at most 3.7\%.
Blocking the answer position lowers it by 16\% for VisionQ-Judge, but only by 4 and 2\% for Gemma-4-E2B and Gemma-4-E4B.
All three judges therefore rely on the image, and the image reaches the answer mainly through the option list rather than through the criterion.

\paragraph{Effect of training.}
In the main text, training on every option order raises accuracy (Sec.~6).
This gain could come from a change in any of the three properties.
VisionQ-Judge is Gemma-4-E4B trained on every option order, so we compare the two models to see which properties training changes.
Position bias decreases: VisionQ-Judge follows the shifted output on 58\% of questions, compared with 37\% for the base model.
Visual grounding becomes stronger: the answer position starts to read the image directly.
Blocking this path lowers the probability of the answer by 15.6\% for VisionQ-Judge, against 1.8\% for the base model.
Grey tiles also lower accuracy more (20 points against 11).
Both models then fall to chance, so this larger drop partly reflects the higher starting accuracy.
Criterion sensitivity does not increase: a different criterion changes fewer answers of VisionQ-Judge than of its base model.
Finally, training does not change how the model perceives images outside judging.
On visual question answering, VisionQ-Judge localises the evidence as well as its base model (AUROC 0.84 for both), and its attention follows the question in the same way.
The accuracy gain in \main{6} is therefore consistent with a lower position bias and a stronger use of the images, rather than with a better use of the criterion.

\paragraph{Limitations.}
Our analysis has four limitations.
First, it covers a single model family, and the knockout results for VisionQ-Judge are based on one training seed.
Whether the findings transfer to other architectures remains to be tested.
Second, attention indicates where a model looks but not necessarily what its answer depends on~\cite{jain2019attention}.
We address this by pairing every attention result with an intervention, either on the input (a different criterion, shifted outputs, grey tiles) or on the attention itself (knockout).
Third, we define sink tokens by attention alone.
These tokens do not show the hidden-state signature that Kang et al.~\cite{kang2025visualsink} use to identify visual attention sinks.
Blocking them changes the answers of all three judges, so we do not treat them as uninformative.
We exclude them only to verify that the overlaps are not driven by them.
Fourth, the main results use the prompt order of the benchmark, in which the image precedes the criterion.
To examine this choice, we placed the criterion before the image or directly before the answer.
For the two base models, the accuracy with the named criterion stays within 3.2 points of the accuracy with a different criterion in all three orders.
For VisionQ-Judge, this difference grows from 1.1 to 5.2 points when the criterion precedes the image.
The weak criterion sensitivity is therefore not explained by the prompt order alone, although the order may influence the trained judge, which we leave to future work.

\FloatBarrier

{
    \small
    \bibliographystyle{ieeenat_fullname}
    \bibliography{references_cvpr}

\begin{thebibliography}{41}
\providecommand{\natexlab}[1]{#1}
\providecommand{\url}[1]{\texttt{#1}}
\expandafter\ifx\csname urlstyle\endcsname\relax
  \providecommand{\doi}[1]{doi: #1}\else
  \providecommand{\doi}{doi: \begingroup \urlstyle{rm}\Url}\fi

\bibitem[Chen et~al.(2024{\natexlab{a}})Chen, Mo, Hou, Wu, Liao, Sun, Yan, and
  Lin]{topiq}
Chaofeng Chen, Jiadi Mo, Jingwen Hou, Haoning Wu, Liang Liao, Wenxiu Sun, Qiong
  Yan, and Weisi Lin.
\newblock {TOPIQ}: A top-down approach from semantics to distortions for image
  quality assessment.
\newblock \emph{IEEE Transactions on Image Processing}, 33:\penalty0
  2404--2418, 2024{\natexlab{a}}.

\bibitem[Chen et~al.(2024{\natexlab{b}})Chen, Chen, Zhang, Wang, Liu, Zhou,
  Zhang, Wan, Zhou, and Sun]{mllmasajudge}
Dongping Chen, Ruoxi Chen, Shilin Zhang, Yaochen Wang, Yinuo Liu, Huichi Zhou,
  Qihui Zhang, Yao Wan, Pan Zhou, and Lichao Sun.
\newblock {MLLM}-as-a-judge: Assessing multimodal {LLM}-as-a-judge with
  vision-language benchmark.
\newblock In \emph{Proceedings of the 41st International Conference on Machine
  Learning}, pages 6562--6595. PMLR, 2024{\natexlab{b}}.

\bibitem[Chen et~al.(2025)Chen, Wen, Du, Zhou, Cui, Han, Weng, Wang, Tong,
  Huang, Chen, Tu, Ye, Zhu, Zhang, Zhou, Zhao, Rafailov, Finn, and
  Yao]{mjbench}
Zhaorun Chen, Zichen Wen, Yichao Du, Yiyang Zhou, Chenhang Cui, Siwei Han, Jen
  Weng, Chaoqi Wang, Zhengwei Tong, Leria Huang, Canyu Chen, Haoqin Tu, Qinghao
  Ye, Zhihong Zhu, Yuqing Zhang, Jiawei Zhou, Zhuokai Zhao, Rafael Rafailov,
  Chelsea Finn, and Huaxiu Yao.
\newblock {MJ-Bench}: Is your multimodal reward model really a good judge for
  text-to-image generation?
\newblock In \emph{Advances in Neural Information Processing Systems (Datasets
  and Benchmarks Track)}, pages 69174--69228, 2025.

\bibitem[{Computer Vision Foundation}(2025)]{cvfopenaccess}
{Computer Vision Foundation}.
\newblock {CVF} open access: {CVPR} 2021 and {CVPR} 2025 papers.
\newblock \url{https://openaccess.thecvf.com}, 2025.

\bibitem[Geva et~al.(2023)Geva, Bastings, Filippova, and
  Globerson]{geva2023dissecting}
Mor Geva, Jasmijn Bastings, Katja Filippova, and Amir Globerson.
\newblock Dissecting recall of factual associations in auto-regressive language
  models.
\newblock In \emph{Proceedings of the Conference on Empirical Methods in
  Natural Language Processing (EMNLP)}, pages 12216--12235, 2023.

\bibitem[Ho et~al.(2026)Ho, Wu, Kumar, Xia, Boudin, Greiner-Petter, and
  Aizawa]{sciclaimeval}
Xanh Ho, Yun-Ang Wu, Sunisth Kumar, Tian~Cheng Xia, Florian Boudin, Andre
  Greiner-Petter, and Akiko Aizawa.
\newblock {SciClaimEval}: Cross-modal claim verification in scientific papers.
\newblock In \emph{Proceedings of the Fifteenth Language Resources and
  Evaluation Conference (LREC)}, pages 11060--11071, 2026.

\bibitem[Hu et~al.(2022)Hu, Shen, Wallis, Allen-Zhu, Li, Wang, Wang, and
  Chen]{lora}
Edward~J. Hu, Yelong Shen, Phillip Wallis, Zeyuan Allen-Zhu, Yuanzhi Li, Shean
  Wang, Lu Wang, and Weizhu Chen.
\newblock {LoRA}: Low-rank adaptation of large language models.
\newblock In \emph{International Conference on Learning Representations}, 2022.

\bibitem[Hu et~al.(2024)Hu, Gao, Hu, Zhang, Chen, Xu, and Wan]{hu2024criteria}
Xinyu Hu, Mingqi Gao, Sen Hu, Yang Zhang, Yicheng Chen, Teng Xu, and Xiaojun
  Wan.
\newblock Are {LLM}-based evaluators confusing {NLG} quality criteria?
\newblock In \emph{Proceedings of the 62nd Annual Meeting of the Association
  for Computational Linguistics (Volume 1: Long Papers)}, pages 9530--9570,
  2024.

\bibitem[Hudson and Manning(2019)]{hudson2019gqa}
Drew~A. Hudson and Christopher~D. Manning.
\newblock {GQA}: A new dataset for real-world visual reasoning and
  compositional question answering.
\newblock In \emph{Proceedings of the IEEE/CVF Conference on Computer Vision
  and Pattern Recognition (CVPR)}, pages 6700--6709, 2019.

\bibitem[Jain and Wallace(2019)]{jain2019attention}
Sarthak Jain and Byron~C. Wallace.
\newblock Attention is not explanation.
\newblock In \emph{Proceedings of the Conference of the North {A}merican
  Chapter of the Association for Computational Linguistics (NAACL)}, pages
  3543--3556, 2019.

\bibitem[Kang et~al.(2025)Kang, Kim, Kim, and Hwang]{kang2025visualsink}
Seil Kang, Jinyeong Kim, Junhyeok Kim, and Seong~Jae Hwang.
\newblock See what you are told: Visual attention sink in large multimodal
  models.
\newblock In \emph{International Conference on Learning Representations
  (ICLR)}, pages 87676--87703, 2025.

\bibitem[Ke et~al.(2021)Ke, Wang, Wang, Milanfar, and Yang]{musiq}
Junjie Ke, Qifei Wang, Yilin Wang, Peyman Milanfar, and Feng Yang.
\newblock {MUSIQ}: Multi-scale image quality transformer.
\newblock In \emph{Proceedings of the IEEE/CVF International Conference on
  Computer Vision (ICCV)}, pages 5148--5157, 2021.

\bibitem[Kil et~al.(2024)Kil, Mai, Lee, Chowdhury, Wang, Cheng, Wang, Liu, and
  Chao]{mllmcompbench}
Jihyung Kil, Zheda Mai, Justin Lee, Arpita Chowdhury, Zihe Wang, Kerrie Cheng,
  Lemeng Wang, Ye Liu, and Wei-Lun Chao.
\newblock {MLLM-CompBench}: A comparative reasoning benchmark for multimodal
  {LLMs}.
\newblock In \emph{Advances in Neural Information Processing Systems (Datasets
  and Benchmarks Track)}, pages 28798--28827, 2024.

\bibitem[Kim et~al.(2026)Kim, Lee, and Park]{vlmsubtlebench}
Minkyu Kim, Sangheon Lee, and Dongmin Park.
\newblock {VLM-SubtleBench}: How far are {VLMs} from human-level subtle
  comparative reasoning?
\newblock In \emph{International Conference on Learning Representations}, pages
  106840--106868, 2026.

\bibitem[Ku et~al.(2024)Ku, Jiang, Wei, Yue, and Chen]{viescore}
Max Ku, Dongfu Jiang, Cong Wei, Xiang Yue, and Wenhu Chen.
\newblock {VIEScore}: Towards explainable metrics for conditional image
  synthesis evaluation.
\newblock In \emph{Proceedings of the 62nd Annual Meeting of the Association
  for Computational Linguistics (Volume 1: Long Papers)}, pages 12268--12290.
  Association for Computational Linguistics, 2024.

\bibitem[Lal et~al.(2025)Lal, Bandham, Hasan, Kashi, Koupaee, and
  Balasubramanian]{musciclaims}
Yash~Kumar Lal, Manikanta Bandham, Mohammad~Saqib Hasan, Apoorva Kashi, Mahnaz
  Koupaee, and Niranjan Balasubramanian.
\newblock {MuSciClaims}: Multimodal scientific claim verification.
\newblock In \emph{Proceedings of the 14th International Joint Conference on
  Natural Language Processing and the 4th Conference of the Asia-Pacific
  Chapter of the Association for Computational Linguistics}, pages 3285--3307,
  2025.

\bibitem[Lee et~al.(2026)Lee, Park, and Im]{mmjudgebias}
Sua Lee, Sanghee Park, and Jinbae Im.
\newblock {MM-JudgeBias}: A benchmark for evaluating compositional biases in
  {MLLM}-as-a-judge.
\newblock In \emph{Proceedings of the 64th Annual Meeting of the Association
  for Computational Linguistics (Volume 1: Long Papers)}, pages 25336--25373.
  Association for Computational Linguistics, 2026.

\bibitem[Li et~al.(2025)Li, Wei, Xie, Yang, Song, Wang, An, Liu, Li, Lin, Kong,
  and Liu]{vlrewardbench}
Lei Li, Yuancheng Wei, Zhihui Xie, Xuqing Yang, Yifan Song, Peiyi Wang, Chenxin
  An, Tianyu Liu, Sujian Li, Bill~Yuchen Lin, Lingpeng Kong, and Qi Liu.
\newblock {VL-RewardBench}: A challenging benchmark for vision-language
  generative reward models.
\newblock In \emph{Proceedings of the IEEE/CVF Conference on Computer Vision
  and Pattern Recognition (CVPR)}, pages 24657--24668, 2025.

\bibitem[Liu et~al.(2026{\natexlab{a}})Liu, Weingord, Mittal, Dungarwal,
  Nandula, Ni, Basu, Chen, Ahmed, Li, Zhang, Goswami, Mukherjee, Kveton,
  Mathur, Dernoncourt, Zhao, Wang, Rossi, Tu, and Du]{mllmjudgeediting}
Runzhou Liu, Hailey Weingord, Sejal Mittal, Prakhar Dungarwal, Anusha Nandula,
  Bo Ni, Samyadeep Basu, Hongjie Chen, Nesreen~K. Ahmed, Li Li, Jiayi Zhang,
  Koustava Goswami, Subhojyoti Mukherjee, Branislav Kveton, Puneet Mathur,
  Franck Dernoncourt, Yue Zhao, Yu Wang, Ryan~A. Rossi, Zhengzhong Tu, and
  Hongru Du.
\newblock Human-aligned {MLLM} judges for fine-grained image editing
  evaluation: A benchmark, framework, and analysis, 2026{\natexlab{a}}.

\bibitem[Liu et~al.(2026{\natexlab{b}})Liu, Chen, Liu, Luo, Tang, Wang, Zeng,
  Dai, Shi, Wei, Lu, Dumoulin, and Tong]{vea}
Zhining Liu, Ziyi Chen, Hui Liu, Chen Luo, Xianfeng Tang, Suhang Wang, Jingying
  Zeng, Zhenwei Dai, Zhan Shi, Tianxin Wei, Hanqing Lu, Benoit Dumoulin, and
  Hanghang Tong.
\newblock Seeing but not believing: Probing the disconnect between visual
  attention and answer correctness in {VLMs}.
\newblock In \emph{International Conference on Learning Representations
  (ICLR)}, pages 72849--72876, 2026{\natexlab{b}}.

\bibitem[Rafailov et~al.(2023)Rafailov, Sharma, Mitchell, Ermon, Manning, and
  Finn]{dpo}
Rafael Rafailov, Archit Sharma, Eric Mitchell, Stefano Ermon, Christopher~D.
  Manning, and Chelsea Finn.
\newblock Direct preference optimization: Your language model is secretly a
  reward model.
\newblock In \emph{Advances in Neural Information Processing Systems}, pages
  53728--53741, 2023.

\bibitem[Reich et~al.(2023)Reich, Putze, and Schultz]{fpvg}
Daniel Reich, Felix Putze, and Tanja Schultz.
\newblock Measuring faithful and plausible visual grounding in {VQA}.
\newblock In \emph{Findings of the Association for Computational Linguistics:
  EMNLP 2023}, pages 3129--3144, 2023.

\bibitem[Shao et~al.(2024)Shao, Qian, Xiao, Song, Zong, Wang, Liu, and
  Li]{shao2024visualcot}
Hao Shao, Shengju Qian, Han Xiao, Guanglu Song, Zhuofan Zong, Letian Wang, Yu
  Liu, and Hongsheng Li.
\newblock Visual {CoT}: Advancing multi-modal language models with a
  comprehensive dataset and benchmark for chain-of-thought reasoning.
\newblock In \emph{Advances in Neural Information Processing Systems (Datasets
  and Benchmarks Track)}, pages 8612--8642, 2024.

\bibitem[Shrestha et~al.(2020)Shrestha, Kafle, and Kanan]{wrongReasons}
Robik Shrestha, Kushal Kafle, and Christopher Kanan.
\newblock A negative case analysis of visual grounding methods for {VQA}.
\newblock In \emph{Proceedings of the 58th Annual Meeting of the Association
  for Computational Linguistics (ACL)}, pages 8172--8181, 2020.

\bibitem[Tamkin et~al.(2024)Tamkin, McCain, Handa, Durmus, Lovitt, Rathi,
  Huang, Mountfield, Hong, Ritchie, Stern, Clarke, Goldberg, Sumers, Mueller,
  McEachen, Mitchell, Carter, Clark, Kaplan, and Ganguli]{tamkin2024clio}
Alex Tamkin, Miles McCain, Kunal Handa, Esin Durmus, Liane Lovitt, Ankur Rathi,
  Saffron Huang, Alfred Mountfield, Jerry Hong, Stuart Ritchie, Michael Stern,
  Brian Clarke, Landon Goldberg, Theodore~R. Sumers, Jared Mueller, William
  McEachen, Wes Mitchell, Shan Carter, Jack Clark, Jared Kaplan, and Deep
  Ganguli.
\newblock Clio: Privacy-preserving insights into real-world {AI} use.
\newblock \emph{arXiv preprint arXiv:2412.13678}, 2024.

\bibitem[Wang et~al.(2023)Wang, Chan, and Loy]{clipiqa}
Jianyi Wang, Kelvin C.~K. Chan, and Chen~Change Loy.
\newblock Exploring {CLIP} for assessing the look and feel of images.
\newblock In \emph{Proceedings of the AAAI Conference on Artificial
  Intelligence}, pages 2555--2563, 2023.

\bibitem[Wu et~al.(2024{\natexlab{a}})Wu, Zhang, Zhang, Chen, Liao, Wang, Li,
  Sun, Yan, Zhai, and Lin]{qbench}
Haoning Wu, Zicheng Zhang, Erli Zhang, Chaofeng Chen, Liang Liao, Annan Wang,
  Chunyi Li, Wenxiu Sun, Qiong Yan, Guangtao Zhai, and Weisi Lin.
\newblock {Q-Bench}: A benchmark for general-purpose foundation models on
  low-level vision.
\newblock In \emph{International Conference on Learning Representations},
  2024{\natexlab{a}}.

\bibitem[Wu et~al.(2024{\natexlab{b}})Wu, Zhu, Zhang, Zhang, Chen, Liao, Li,
  Wang, Sun, Yan, Liu, Zhai, Wang, and Lin]{coinstruct}
Haoning Wu, Hanwei Zhu, Zicheng Zhang, Erli Zhang, Chaofeng Chen, Liang Liao,
  Chunyi Li, Annan Wang, Wenxiu Sun, Qiong Yan, Xiaohong Liu, Guangtao Zhai,
  Shiqi Wang, and Weisi Lin.
\newblock Towards open-ended visual quality comparison.
\newblock In \emph{Computer Vision -- ECCV 2024}, pages 360--377. Springer,
  2024{\natexlab{b}}.

\bibitem[Wu et~al.(2025)Wu, Zou, Liang, Zhang, and Ma]{visualqualityr1}
Tianhe Wu, Jian Zou, Jie Liang, Lei Zhang, and Kede Ma.
\newblock {VisualQuality-R1}: Reasoning-induced image quality assessment via
  reinforcement learning to rank.
\newblock In \emph{Advances in Neural Information Processing Systems}, pages
  88167--88190, 2025.

\bibitem[Wu et~al.(2023)Wu, Hao, Sun, Chen, Zhu, Zhao, and Li]{hpsv2}
Xiaoshi Wu, Yiming Hao, Keqiang Sun, Yixiong Chen, Feng Zhu, Rui Zhao, and
  Hongsheng Li.
\newblock Human preference score v2: A solid benchmark for evaluating human
  preferences of text-to-image synthesis.
\newblock \emph{arXiv preprint arXiv:2306.09341}, 2023.

\bibitem[Xie et~al.(2026)Xie, Zhang, Wu, Lu, Zhang, Yu, Wang, Hong, Liu, Wu,
  and Luo]{visjudge}
Yupeng Xie, Zhiyang Zhang, Yifan Wu, Sirong Lu, Jiayi Zhang, Zhaoyang Yu,
  Jinlin Wang, Sirui Hong, Bang Liu, Chenglin Wu, and Yuyu Luo.
\newblock {VisJudge-Bench}: Aesthetics and quality assessment of
  visualizations.
\newblock In \emph{International Conference on Learning Representations}, pages
  85976--86037, 2026.

\bibitem[Xiong et~al.(2025)Xiong, Wang, Guo, Ye, Fan, Gu, Huang, and
  Li]{llavacritic}
Tianyi Xiong, Xiyao Wang, Dong Guo, Qinghao Ye, Haoqi Fan, Quanquan Gu, Heng
  Huang, and Chunyuan Li.
\newblock {LLaVA-Critic}: Learning to evaluate multimodal models.
\newblock In \emph{Proceedings of the IEEE/CVF Conference on Computer Vision
  and Pattern Recognition (CVPR)}, pages 13618--13628, 2025.

\bibitem[Xu et~al.(2023)Xu, Liu, Wu, Tong, Li, Ding, Tang, and
  Dong]{imagereward}
Jiazheng Xu, Xiao Liu, Yuchen Wu, Yuxuan Tong, Qinkai Li, Ming Ding, Jie Tang,
  and Yuxiao Dong.
\newblock {ImageReward}: Learning and evaluating human preferences for
  text-to-image generation.
\newblock In \emph{Advances in Neural Information Processing Systems}, pages
  15903--15935, 2023.

\bibitem[Ye et~al.(2025)Ye, Li, Li, Ai, Zhou, Shen, Yan, and Liu]{conj}
Ziyi Ye, Xiangsheng Li, Qiuchi Li, Qingyao Ai, Yujia Zhou, Wei Shen, Dong Yan,
  and Yiqun Liu.
\newblock Learning {LLM}-as-a-judge for preference alignment.
\newblock In \emph{International Conference on Learning Representations}, pages
  3537--3564, 2025.

\bibitem[Zeiler and Fergus(2014)]{zeiler2014visualizing}
Matthew~D. Zeiler and Rob Fergus.
\newblock Visualizing and understanding convolutional networks.
\newblock In \emph{European Conference on Computer Vision (ECCV)}, pages
  818--833, 2014.

\bibitem[Zhang et~al.(2025{\natexlab{a}})Zhang, Khayatkhoei, Chhikara, and
  Ilievski]{whereToLook}
Jiarui Zhang, Mahyar Khayatkhoei, Prateek Chhikara, and Filip Ilievski.
\newblock {MLLMs} know where to look: Training-free perception of small visual
  details with multimodal {LLMs}.
\newblock In \emph{International Conference on Learning Representations
  (ICLR)}, pages 68194--68213, 2025{\natexlab{a}}.

\bibitem[Zhang et~al.(2024)Zhang, Wu, Zhang, Zhai, and Lin]{qbenchplus}
Zicheng Zhang, Haoning Wu, Erli Zhang, Guangtao Zhai, and Weisi Lin.
\newblock {Q-Bench+}: A benchmark for multi-modal foundation models on
  low-level vision from single images to pairs.
\newblock \emph{IEEE Transactions on Pattern Analysis and Machine
  Intelligence}, 46\penalty0 (12):\penalty0 10404--10418, 2024.

\bibitem[Zhang et~al.(2025{\natexlab{b}})Zhang, Wu, Li, Zhou, Sun, Min, Chen,
  Liu, Lin, and Zhai]{abench}
Zicheng Zhang, Haoning Wu, Chunyi Li, Yingjie Zhou, Wei Sun, Xiongkuo Min,
  Zijian Chen, Xiaohong Liu, Weisi Lin, and Guangtao Zhai.
\newblock {A-Bench}: Are {LMMs} masters at evaluating {AI}-generated images?
\newblock In \emph{International Conference on Learning Representations}, pages
  84824--84845, 2025{\natexlab{b}}.

\bibitem[Zhang et~al.(2025{\natexlab{c}})Zhang, Yadav, Han, and
  Shutova]{crossmodalflow}
Zhi Zhang, Srishti Yadav, Fengze Han, and Ekaterina Shutova.
\newblock Cross-modal information flow in multimodal large language models.
\newblock In \emph{Proceedings of the IEEE/CVF Conference on Computer Vision
  and Pattern Recognition (CVPR)}, pages 19781--19791, 2025{\natexlab{c}}.

\bibitem[Zheng et~al.(2024)Zheng, Zhou, Meng, Zhou, and Huang]{wang2023large}
Chujie Zheng, Hao Zhou, Fandong Meng, Jie Zhou, and Minlie Huang.
\newblock Large language models are not robust multiple choice selectors.
\newblock In \emph{International Conference on Learning Representations}, pages
  19426--19454, 2024.

\bibitem[Zheng et~al.(2023)Zheng, Chiang, Sheng, Zhuang, Wu, Zhuang, Lin, Li,
  Li, Xing, Zhang, Gonzalez, and Stoica]{mtbench}
Lianmin Zheng, Wei-Lin Chiang, Ying Sheng, Siyuan Zhuang, Zhanghao Wu, Yonghao
  Zhuang, Zi Lin, Zhuohan Li, Dacheng Li, Eric~P. Xing, Hao Zhang, Joseph~E.
  Gonzalez, and Ion Stoica.
\newblock Judging {LLM}-as-a-judge with {MT-Bench} and chatbot arena.
\newblock In \emph{Advances in Neural Information Processing Systems}, 2023.

\end{thebibliography}
}

\end{document}